\documentclass[letterpaper]{article}
\usepackage{times}
\usepackage{helvet}
\usepackage{courier}

\usepackage{mathrsfs}
\usepackage{dsfont}
\usepackage{amsmath}
\usepackage{amsthm}
\usepackage{amssymb}

\newtheorem{thm}{Theorem}
\newtheorem{cor}{Corollary}
\newtheorem{lem}{Lemma}
\newtheorem{prop}{Proposition}
\theoremstyle{definition}

\theoremstyle{remark}
\newtheorem{rem}{Remark}
\theoremstyle{definition}
\newtheorem{exm}{Example}

\begin{document}
%
\title{
Expressiveness of Logic Programs under General Stable Model Semantics
}
\author{Heng Zhang \and Yan Zhang\\
Artificial Intelligence Research Group\\
School of Computing, Engineering and Mathematics\\ University of Western Sydney, Australia
}
\maketitle
\begin{abstract}
\begin{quote}
The stable model semantics had been recently generalized to non-Herbrand structures by several works, which provides a unified framework and solid logical foundations for answer set programming. This paper focuses on the expressiveness of normal and disjunctive programs under the general stable model semantics. A translation from disjunctive programs to normal programs is proposed for infinite structures. Over finite structures, some disjunctive programs are proved to be intranslatable to normal programs if the arities of auxiliary predicates and functions are bounded in a certain way. The equivalence of the expressiveness of normal programs and disjunctive programs over arbitrary structures is also shown to coincide with that over finite structures, and coincide with whether NP is closed under complement. Moreover, to capture the exact expressiveness, some intertranslatability results between logic program classes and fragments of second-order logic are obtained.
\end{quote}
\end{abstract}

\section{Introduction}

Logic programming with default negation is an elegant and efficient formalism for knowledge representation, which incorporates the abilities of classical logic, inductive definition and commonsense reasoning. Nowadays, the most popular semantics for this formalism is the stable model semantics proposed by~\cite{GL:88}. Logic programming based on this semantics, which is known as answer-set programming, has then emerged as a flourishing paradigm for declarative programming in the last fifteen years.

The original stable model semantics focuses only on Herbrand structures in which the unique name assumption is made. For a certain class of applications, this assumption will simplify the representation. However, there are many applications where the knowledge can be more naturally represented over non-Herbrand structures, including arithmetical structures and other metafinite structures. To overcome this limit, the general stable model semantics, which generalizes the original semantics to arbitrary structures, was then proposed by~\cite{FLL:11} via second-order logic, by~\cite{LZ:11} via circumscription, and by~\cite{PV:05} via G\"{o}del's 3-valued logic,
which provides us a unified framework for answer set programming, armed with powerful tools from classical logic.

%
%
%
%

The main goal of this work is to identify the expressiveness of logic programs, which is one of the central topics in knowledge representation and reasoning. We will focus on two important classes of logic programs -- normal programs and disjunctive programs. Over Herbrand structures, the expressiveness of logic programs under the query equivalence has been thoroughly studied in the last three decades. An excellent survey for these works can be found in~\cite{DEGV:01}. Our task is quite different. On the one hand, we will work on the general stable model semantics so that non-Herbrand structures will be considered. On the other hand, instead of considering query equivalence, the expressiveness in our work will be based on model equivalence. This setting is important since answer set solvers are usually used to generate models. The model equivalence implies the query equivalence, but the converse is in general not true.

We also hope this work contributing to the effective implementation of answer set solvers. Translating logic programs into classical logics is a usual approach to implement answer set solvers, e.g.,~\cite{LZ:04,LM:04}. In this work, we are interested in translating normal programs to first-order sentences so that the state-of-the-art SMT solvers can be used for answer set solving. As the arity of auxiliary symbol is the most important factor to introduce nondeterminism~\cite{Imme:99}, we will try to find translations with small arities of auxiliary symbols.  

Our contribution in this paper is fourfold. Firstly, we show that, over infinite structures, every disjunctive program can be equivalently translated to a normal one. Secondly, we prove that, if finite structures are focused, for each integer $n$ greater than $1$ there is a disjunctive program with intensional predicates of arities less than $n$ that cannot be equivalently translated to any normal program with auxiliary predicates of arities less than $2n$.  Thirdly, we show that disjunctive and normal programs are of the same expressiveness over arbitrary structures if and only if they are of the same expressiveness over finite structures, if and only if the complexity class NP is closed under complement. Lastly, to understand the exact expressiveness of logic programs, we also prove that the intertranslatability holds between some classes of logic programs and some fragments of second-order logic.

\section{Preliminaries}

{\em Vocabularies} are assumed to be sets of predicate constants
and function constants.
Logical symbols are as usual, including a countable set of
predicate variables and a countable set of function variables. Every constant or variable is equipped with a natural number, its {\em arity}. Nullary function constants and variables are called {\em individual constants} and {\em variables} respectively. Nullary predicate constants are called {\em propositions}. Sometimes, we do not distinguish between constants and variables, and simply call them {\em predicates} or {\em functions}
if no confusion occurs.
Atoms, formulas, sentences and theories of a vocabulary $\upsilon$ (or shortly,
$\upsilon$-atoms, $\upsilon$-formulas, $\upsilon$-sentences and $\upsilon$-theories) are built
from $\upsilon$, equality, variables, connectives and quantifiers in a standard way. Every {\em positive clause} of $\upsilon$ is a finite disjunction of $\upsilon$-atoms.
Given a sentence $\varphi$ and a theory $\Sigma$, let $\upsilon(\varphi)$ and $\upsilon(\Sigma)$ denote the sets of constants occurring in $\varphi$ and $\Sigma$ respectively.

Assuming $\mathsf{Q}$ to be $\forall$ or $\exists$, let $\mathsf{Q}\tau$ and $\mathsf{Q}\bar{x}$ denote the quantifier blocks $\mathsf{Q}X_1\cdots\mathsf{Q}X_n$ and $\mathsf{Q}x_1\cdots\mathsf{Q}x_m$ respectively if $\tau$ is the finite set $\{X_1,\dots,X_n\}$, $\bar{x}=x_1\cdots x_m$, $X_i$ and $x_i$ are predicate/function and individual variables respectively.
Let $\Sigma${\small$^{1\textsc{f}}_{n,k}$} be the class of sentences of form
$
\mathsf{Q}_1\tau_1\cdots\mathsf{Q}_n\tau_n\varphi
$,
where $\mathsf{Q}_i$ is $\exists$ if $i$ is odd, otherwise it is $\forall$; $\tau_i$ is a finite set of variables of arities $\le k$; and no second-order quantifier 
appears in $\varphi$. Let $\Sigma${\small$^{1}_{n,k}$} denote the class defined as the same as $\Sigma${\small$^{1\textsc{f}}_{n,k}$} except no function variable allowed in any $\tau_i$. Let $\Sigma${\small$^{1\textsc{f}}_{n}$} (respectively, $\Sigma${\small$^{1}_{n}$}) be the union of $\Sigma${\small$^{1\textsc{f}}_{n,k}$} (respectively, $\Sigma${\small$^{1}_{n,k}$}) for all $k\ge 0$. Given a class $\Sigma$ defined as above, let $\Sigma[\forall^{\ast}\exists^{\ast}]$ (respectively, $\Sigma[\forall^{\ast}]$) be the class of sentences in $\Sigma$ with first-order part of form $\forall\bar{x}\exists\bar{y}\vartheta$ (respectively, $\forall\bar{x}\vartheta$), where $\bar{x}$ and $\bar{y}$ are tuples of individual variables, and $\vartheta$ quantifier-free.

%
%


Every {\em structure} $\mathds{A}$ of $\upsilon$ (or
shortly, {\em $\upsilon$-structure} $\mathds{A}$) is accompanied by
a nonempty set $A$, the {\em domain} of $\mathds{A}$, and
interprets each $n$-ary predicate constant $P$ in $\upsilon$ as an
$n$-ary relation $P^{\mathds{A}}$ on $A$, and interprets each
$n$-ary function constant $f$ in $\upsilon$ as an $n$-ary function
$f^{\mathds{A}}$ on $A$. A structure is {\em finite} if its domain is finite; otherwise it is {\em infinite}. Let $\mathsf{FIN}$ denote the class of finite structures, and let $\mathsf{INF}$ denote the class of infinite structures.
A {\em restriction} of
a structure $\mathds{A}$ to a vocabulary $\sigma$ is the structure obtained
from $\mathds{A}$ by discarding all interpretations for constants
not in $\sigma$. Given a vocabulary $\upsilon\supset\sigma$ and a
$\sigma$-structure $\mathds{B}$,
every {\em $\upsilon$-expansion} of $\mathds{B}$ is a structure $\mathds{A}$ of $\upsilon$ such that
$\mathds{B}$ is a restriction of $\mathds{A}$ to $\sigma$. Given a structure $\mathds{A}$ and a set $\tau$ of predicates,
let $\textsc{Ins}(\mathds{A},\tau)$ be the set of {\em ground atoms} $P(\bar{a})$ for all $\bar{a}\in P^{\mathds{A}}$ and all $P$ in $\tau$.

Every {\em assignment} in a structure
$\mathds{A}$ is a function that maps each individual
variable to an element of $A$ and maps each predicate (respectively, function)
variable to a relation (respectively, function) on $A$ of the same arity. Given a formula $\varphi$
and an assignment $\alpha$ in $\mathds{A}$, write
$\mathds{A}\models\varphi[\alpha]$ if $\alpha$ {\em satisfies}
$\varphi$ in $\mathds{A}$ in the standard way. In particular, if
$\varphi$ is a sentence, simply write $\mathds{A}\models\varphi$, and say
$\mathds{A}$ is a {\em model} of $\varphi$, or in other
words, $\varphi$ {\em is true} in $\mathds{A}$. Given formulas $\varphi,\psi$ and a class $\mathcal{C}$ of structures, we say $\varphi$ is {\em equivalent} to $\psi$ over $\mathcal{C}$, or write $\varphi\equiv_{\mathcal{C}}\psi$ for short, if for every $\mathds{A}$ in $\mathcal{C}$ and every assignment $\alpha$ in $\mathds{A}$, $\alpha$ satisfies $\varphi$ in $\mathds{A}$ if and only if $\alpha$ satisfies $\psi$ in $\mathds{A}$. Given a quantifier-free formula $\varphi$
and an assignment $\alpha$ in $\mathds{A}$, let $\varphi[\alpha]$ denote the {\em ground formula} obtained from $\varphi$ by substituting $a$ for $t$ whenever $a=\alpha(t)$ and $t$ is a term.
%



A class of structures is also called a {\em property}. Let $\mathcal{C}$ and $\mathcal{D}$ be two properties. We say $\mathcal{D}$ is {\em defined} by a sentence $\varphi$ over $\mathcal{C}$, or equivalently, $\varphi$ {\em defines} $\mathcal{D}$ over $\mathcal{C}$, if each structure of $\mathcal{C}$ is in $\mathcal{D}$ if and only if it is a model of $\varphi$; $\mathcal{D}$ is {\em definable} in a class $\Sigma$ of sentences over $\mathcal{C}$ if there is a sentence in $\Sigma$ that defines $\mathcal{D}$ over $\mathcal{C}$. Given two classes $\Sigma,\Lambda$ of sentences, we write $\Sigma\leq_{\mathcal{C}}\Lambda$ if each property definable in $\Sigma$ over $\mathcal{C}$ is also definable in $\Lambda$ over $\mathcal{C}$; we write $\Sigma\simeq_{\mathcal{C}}\Lambda$ if both $\Sigma\leq_{\mathcal{C}}\Lambda$ and $\Lambda\leq_{\mathcal{C}}\Sigma$ hold. In particular, if $\mathcal{C}$ is the class of arbitrary structures, the subscript $\mathcal{C}$ can be dropped.

\subsection{Logic Programs and Stable Models}

Every {\em disjunctive program} is a set of {\em rules} of the form
\begin{equation}\label{eqn:rule}
\zeta_1\wedge\cdots\wedge\zeta_{m}\rightarrow\zeta_{m+1}\vee\cdots\vee\zeta_{n}
\end{equation}
where $0\leq m\leq n$ and $n>0$; $\zeta_i$ is an atom not involving the equality if $m<i\leq n$; $\zeta_j$ is a {\em literal}, i.e. an atom or the negation of an atom, if $1\leq j\leq m$. Given a rule, the disjunctive part is called its {\em head}, and the conjunctive part is called its {\em body}. Given a disjunctive program $\Pi$, a predicate is called {\em intensional} (w.r.t. $\Pi$) if it appears in the head of some rule in $\Pi$; a formula is called {\em intensional} (w.r.t. $\Pi$) if it does not involve any non-intensional predicate. Let $\upsilon(\Pi)$ be the set of predicates and functions appearing in $\Pi$.

Let $\Pi$ be a disjunctive program. Then $\Pi$ is called {\em normal} if the head of each rule contains at most one atom, $\Pi$ is {\em plain} if the negation of any intensional atom does not appear in the body of any rule,
$\Pi$ is {\em propositional} if it does not involve any predicate of positive arity, and $\Pi$ is {\em finite} if it contains only a finite set of rules. In particular, unless mentioned otherwise, a disjunctive program is always {\em assumed to be finite}.

Given any disjunctive program $\Pi$, let $\mathrm{SM}(\Pi)$ denote the second-order sentence
$\varphi\wedge\forall\tau^{\ast}(\tau^{\ast}<\tau\rightarrow\neg\varphi^{\ast})$,
where $\tau$ is the set of intensional predicates; $\tau^{\ast}$ is the set of predicate variables $P^{\ast}$ for all predicates $P$ in $\tau$; $\tau^{\ast}<\tau$ is the formula
$\wedge_{P\in\tau}\forall\bar{x}(P^{\ast}(\bar{x})\rightarrow P(\bar{x}))\wedge\neg\wedge_{P\in\tau}\forall\bar{x}(P(\bar{x})\rightarrow P^{\ast}(\bar{x}))$;
$\varphi$ is the conjunction of all the sentences $\gamma${\small$_{\forall}$} such that $\gamma$ is a rule in $\Pi$ and $\gamma${\small$_{\forall}$} is the first-order universal closure of $\gamma$; $\varphi^{\ast}$ is the formula obtained from $\varphi$ by substituting $P^{\ast}$ for all positive occurrences of $P$ in the head or in the body of each rule if $P$ is in $\tau$. (So, all negations in intensional literals are default negations.) A structure $\mathds{A}$ is called a {\em stable model} of $\Pi$ if it satisfies $\mathrm{SM}(\Pi)$. For more details about this transformation, please refer to~\cite{FLL:11}.

Given two properties $\mathcal{C}$ and $\mathcal{D}$, we say $\mathcal{D}$ is {\em defined} by a disjunctive program $\Pi$ over $\mathcal{C}$ via the set $\tau$ of {\em auxiliary constants} if the formula $\exists\tau\mathrm{SM}(\Pi)$ defines $\mathcal{D}$ over $\mathcal{C}$, where $\tau$ is a set of predicates and functions occurring in $\Pi$. Given $n\geq 0$, let $\mathrm{DLP}_n$ (respectively, $\mathrm{DLP}_n^{\textsc{f}}$) be the class of sentences $\exists\tau\mathrm{SM}(\Pi)$ for all disjunctive programs $\Pi$ and all finite sets $\tau$ of predicate (respectively, predicate and function) constants of arities $\leq n$. Let $\mathrm{DLP}$ (respectively, $\mathrm{DLP}^{\textsc{f}}$) be the union of $\mathrm{DLP}_n$ (respectively, $\mathrm{DLP}_n^{\textsc{f}}$) for all $n\geq 0$. In above definitions, if $\Pi$ is restricted to be normal, we then obtain the notations $\mathrm{NLP}_n,\mathrm{NLP}_n^{\textsc{f}},\mathrm{NLP}$ and $\mathrm{NLP}^{\textsc{f}}$ respectively.


Given a rule $\gamma$, let $\gamma${\small$^-_{\textsc{B}}$} be the set of conjuncts in the body of $\gamma$ in which no intensional predicate positively occurs, and let $\gamma${\small$^+$} be the rule obtained from $\gamma$ by removing all literals in $\gamma${\small$^-_{\textsc{B}}$}. Given a disjunctive program $\Pi$ and a structure $\mathds{A}$, let $\Pi${\small$^{\mathds{A}}$} be the set of rules $\gamma${\small$^{+}[\alpha]$} for all assignments $\alpha$ in $\mathds{A}$ and all rules $\gamma$ in $\Pi$ such that $\alpha$ satisfies $\gamma${\small$^-_{\textsc{b}}$} in $\mathds{A}$. Now, $\Pi^{\mathds{A}}$ can be regarded as a propositional program where each ground atom as a proposition. This procedure is called the {\em first-order Gelfond-Lifschitz reduction} due to the following result:

\begin{prop}[\cite{ZZ:13}, Proposition 4]\label{prop:fo2prop}
Let $\Pi$ be a disjunctive program and $\tau$ the set of intensional predicates. Then an $\upsilon(\Pi)$-structure $\mathds{A}$ is a stable model of $\Pi$ iff $\textsc{Ins}(\mathds{A},\tau)$ is a minimal (w.r.t. the set inclusion) model of $\Pi^{\mathds{A}}$.
\end{prop}
%

\subsection{Progression Semantics}

In this subsection, we review a progression semantics proposed by~\cite{ZZ:13}, which generalizes the fixed point semantics of~\cite{LMR:92} to logic programming with default negation over arbitrary structures. For convenience, two positive clauses consisting of the same set of atoms will be regarded as the same.

%

\smallskip
Let $\Pi$ be a propositional, possibly infinite and plain disjunctive program. Let $\textsc{pc}(\upsilon(\Pi))$ denote the set of positive clauses of $\upsilon(\Pi)$ and let $\Lambda\subseteq\textsc{pc}(\upsilon(\Pi))$. Define $\Gamma_{\Pi}(\Lambda)$ to be
{\small
\begin{equation*}
\left\{H\vee C_1\vee\cdots\vee C_k\left|
\begin{aligned}
k\ge 0&\,\,\&\,\,H,C_1,\dots,C_k\in\textsc{pc}(\upsilon(\Pi))\\
&\,\,\&\,\,\exists p_1,\dots,p_k\in\upsilon(\Pi)\text{ s.t. } \\
&
\,\left[
\begin{aligned}
&p_1\wedge\cdots\wedge p_k\rightarrow H\in\Pi\,\,\&\\
&C_1\vee p_1,\dots,C_k\vee p_k\in\Lambda
\end{aligned}
\right]
\end{aligned}
\right.
\right\}.
\end{equation*}
}

\noindent It is clear that $\Gamma_{\Pi}$ is a monotone operator on $\textsc{pc}(\upsilon(\Pi))$.

Now, a progression operator for first-order programs can then be defined via the first-order Gelfond-Lifschitz reduction. Given a disjunctive program $\Pi$ and an $\upsilon(\Pi)$-structure $\mathds{A}$, define  $\Gamma${\small$^{\mathds{A}}_{\Pi}$} to be the operator $\Gamma_{\Pi^{\mathds{A}}}$; let $\Gamma${\small$^{\mathds{A}}_{\Pi}$}\,$\uparrow_0$ denote the empty set, and let $\Gamma${\small$^{\mathds{A}}_{\Pi}$}\,$\uparrow_n$ denote $\Gamma${\small$^{\mathds{A}}_{\Pi}($}$\Gamma${\small$^{\mathds{A}}_{\Pi}$}\,$\uparrow_{n-1})$ for all $n>0$; finally, let $\Gamma${\small$^{\mathds{A}}_{\Pi}$}\,$\uparrow_{\omega}$ be the union of $\Gamma${\small$^{\mathds{A}}_{\Pi}$}\,$\uparrow_n$ for all $n\geq 0$. To illustrate the definitions, a simple example is given as follows.

\begin{exm}
Let $\Pi$ be the logic program consisting of rules
\begin{equation*}
S(x)\vee T(x)\quad\text{and}\quad T(x)\wedge E(x,y)\rightarrow T(y).
\end{equation*}
Let $\upsilon$ be $\{E\}$ and $\mathds{A}$ a structure of $\upsilon$. Then, for $n>0$, $\Gamma${\small$^{\mathds{A}}_{\Pi}$}\,$\uparrow_n$ is the set of clauses $S(a)\vee T(b)$ such that $a,b\in A$ and there exists a path from $a$ to $b$ via $E$ of length less than $n$.\hfill$\Box$
\end{exm}

The following proposition shows that the general stable model semantics can be defined by the progression operator.

\begin{prop}[\cite{ZZ:13}, Theorem 1]\label{prop:fxp2sm}
Let $\Pi$ be a disjunctive program, $\tau$ the set of intensional predicates of $\Pi$, and $\mathds{A}$ a structure of $\upsilon(\Pi)$. Then
$\mathds{A}$ is a stable model of $\Pi$ iff $\textsc{Ins}(\mathds{A},\tau)$ is a minimal model of $\Gamma${\small$^{\mathds{A}}_{\Pi}$}\,$\uparrow_{\omega}$.
\end{prop}

\begin{rem}
In Proposition \ref{prop:fxp2sm}, it is clear that, if $\Pi$ is normal, $\mathds{A}$ is a stable model of $\Pi$ if and only if $\textsc{Ins}(\mathds{A},\tau)=\Gamma${\small$^{\mathds{A}}_{\Pi}$}\,$\uparrow_{\omega}$.
\end{rem}

%
%
%
%
%
%
%
%
%

\section{Infinite Structures}

This section will focus on the expressiveness of logic programs over infinite structures. We first propose a translation that reduces each disjunctive program to a normal program over infinite structures. The main idea is to encode grounded positive clauses by elements in the intended domain. With the encoding, we then simulate the progression of the given disjunctive program by the progression of a normal program.

\smallskip
We first show how to encode a positive clause by an element.
Let $A$ be an infinite set. Each {\em encoding function} on $A$ is defined to be an injective function from $A\times A$ into $A$. Let $\textsl{enc}$ be an encoding function on $A$ and $c$ an element in $A$ such that $\textsl{enc}(a,b)\neq c$ for all elements $a,b\in A$. To simplify the statement, let $\textsl{enc}(a_1,\dots,a_k;c)$ denote the expression
\begin{equation}
\textsl{enc}((\cdots(\textsl{enc}(c,a_1),a_2),\cdots),a_k)
\end{equation} for any $k\geq 0$ and any set of elements $a_1,\dots,a_k\in A$. In the above expression, the special element $c$ is used as a flag to indicate that the encoded tuple will be started after $c$, and is then called the {\em encoding flag} of this encoding.

Let $A^{\ast}$ denote the set of finite tuples of elements in $A$ and $\textsl{enc}[A,c]$ the set of elements $\textsl{enc}(\bar{a};c)$ for all tuples $\bar{a}$ in $A^{\ast}$. The {\em merging function} $\textsl{mrg}$ on $A$ related to $\textsl{enc}$ and $c$ is the function from $\textsl{enc}[A,c]\times\textsl{enc}[A,c]$ into $\textsl{enc}[A,c]$ such that
\begin{equation}
\textsl{mrg}(\textsl{enc}(\bar{a};c),\textsl{enc}(\bar{b};c))=\textsl{enc}(\bar{a},\bar{b};c)
\end{equation}
for all tuples $\bar{a}$ and $\bar{b}$ in $A^{\ast}$. Again, to simplify the statement, we let $\textsl{mrg}(a_1,\dots,a_k)$ be short for the expression
\begin{equation}
\textsl{mrg}((\cdots(\textsl{mrg}(a_1,a_2),a_3),\cdots),a_k)
\end{equation}
if all encoding flags of $a_1,\dots,a_k$ are the same.
It is clear that the merging function is unique if $\textsl{enc}$ and $c$ are fixed.

\begin{exm}
Let $\mathbb{Z}^+$, the set of all positive integers, be the domain that we will focus, and let $P_1,P_2,P_3$ be three  predicates of arities $2,3,1$ respectively. Next, we show how to encode ground positive clauses by integers in $\mathbb{Z}^+$.

Let $e$ be a function from $\mathbb{Z}^+\times\mathbb{Z}^+$ into $\mathbb{Z}^+$ such that
\begin{equation}
e(m,n)=2^{m}+3^{n}\text{ for all }m,n\in\mathbb{Z}^+.
\end{equation}
It is easy to check that $e$ is an encoding function on $\mathbb{Z}^+$, and integers $1,2,3,4$ are not in the range of $e$. For $1\le i\le 3$, let $i$ be the encoding flag for the encodings of atoms built from $P_i$. Then the grounded atom $P_2(1,3,5)$ can be encoded by
\begin{equation}
e(1,3,5;2)=e(e(e(2,1),3),5)=2^{155}+3^5.
\end{equation}
Let $4$ be the encoding flag for encodings of positive clauses. Then the positive clause $P_2(1,3,5)\vee P_3(2)\vee P_1(2,4)$ can be encoded by $e(e(1,3,5;2),e(2;3),e(2,4;1);4)$.\hfill$\Box$
\end{exm}

In classical logic, two positive clauses are equivalent if and only if they contain the same set of atoms. Assume that $c$ is the encoding flag for encodings of positive clauses and $\textsl{enc}$ is the encoding function. To capture the equivalence between two positive clauses, some encoding predicates related to $\textsl{enc}$ and $c$ are needed. We define them as follows:
\begin{eqnarray}
\label{eqn:defn_in}\!\!\!\!\textsl{in}\!\!\!\!&=&\!\!\!\!\{(\textsl{enc}(\bar{a};c),b)\mid\bar{a}\in A^{\ast}\wedge b\in[\bar{a}]\},\\
\label{eqn:defn_subc}\!\!\!\!\textsl{subc}\!\!\!\!&=&\!\!\!\!\{(\textsl{enc}(\bar{a};c),\textsl{enc}(\bar{b};c))\mid\bar{a},\bar{b}\in A^{\ast}\wedge[\bar{a}]\subseteq[\bar{b}]\},\\
\label{eqn:defn_equ}\!\!\!\!\textsl{equ}\!\!\!\!&=&\!\!\!\!\{(\textsl{enc}(\bar{a};c),\textsl{enc}(\bar{b};c))\mid\bar{a},\bar{b}\in A^{\ast}\wedge[\bar{a}]=[\bar{b}]\},
\end{eqnarray}
where $[\bar{a}],[\bar{b}]$ are the sets of elements in $\bar{a},\bar{b}$ respectively. 
Intuitively, $\textsl{in}(a,b)$ expresses that the atom encoded by $b$ appears in the positive clause encoded by $a$; $\textsl{subc}(a,b)$ expresses that the positive clause encoded by $a$ is a subclause of that encoded by $b$; and $\textsl{equ}(a,b)$ expresses that the positive clauses encoded by $a$ and $b$ respectively are equivalent.

\smallskip
With the above method for encoding, we can then define a translation.
Let $\Pi$ be a disjunctive program. We first construct a class of normal programs related to $\Pi$ as follows:

\smallskip

1. Let $C_{\Pi}$ denote the set consisting of an individual constant $c_{P}$ for each predicate constant $P$ that occurs in $\Pi$, and of an individual constant $c_{\epsilon}$, where $c_{\epsilon}$ will be interpreted as the encoding flag for positive clauses, and $c_P$ as the encoding flag for atoms built from $P$. Let $\Pi_1$ consist of the rule
\begin{eqnarray}
\label{eqn:pi1_1}\textsc{enc}(x,y,c)&\rightarrow&\bot
\end{eqnarray}
for each individual constant $c\in C_{\Pi}$, and the following rules:
\begin{eqnarray}
\label{eqn:pi1_2}\!\!\!\!\!\!\!\!\!\!\neg\underline{\textsc{enc}}(x,y,z)\!\!\!\!&\rightarrow&\!\!\!\!\textsc{enc}(x,y,z)\\
\label{eqn:pi1_3}\!\!\!\!\!\!\!\!\!\!\neg\textsc{enc}(x,y,z)\!\!\!\!&\rightarrow&\!\!\!\!\underline{\textsc{enc}}(x,y,z)\\
\label{eqn:pi1_4}\!\!\!\!\!\!\!\!\!\!\textsc{enc}(x,y,z)\wedge\textsc{enc}(u,v,z)\wedge\neg x=u\!\!\!\!&\rightarrow&\!\!\!\!\bot\\
\label{eqn:pi1_5}\!\!\!\!\!\!\!\!\!\!\textsc{enc}(x,y,z)\wedge\textsc{enc}(u,v,z)\wedge\neg y=v\!\!\!\!&\rightarrow&\!\!\!\!\bot\\
\label{eqn:pi1_6}\!\!\!\!\!\!\!\!\!\!\textsc{enc}(x,y,z)\!\!\!\!&\rightarrow&\!\!\!\!\textsc{ok}_e(x,y)\\
\label{eqn:pi1_7}\!\!\!\!\!\!\!\!\!\!\neg\textsc{ok}_e(x,y)\!\!\!\!&\rightarrow&\!\!\!\!\textsc{ok}_e(x,y)\\
\label{eqn:pi1_8}\!\!\!\!\!\!\!\!\!\!\textsc{enc}(x,y,z)\wedge\textsc{enc}(x,y,u)\wedge\neg z=u\!\!\!\!&\rightarrow&\!\!\!\!\bot
\end{eqnarray}
Informally, rules (\ref{eqn:pi1_6})--(\ref{eqn:pi1_8}) describe that $\textsc{enc}$ is the graph of a function; rules (\ref{eqn:pi1_4})--(\ref{eqn:pi1_5}) describe that $\textsc{enc}$ is injective. Thus, $\textsc{enc}$ should be the graph of an encoding function. In addition, rule (\ref{eqn:pi1_1}) assures that $c$ is not in the range of $\textsc{enc}$.

\smallskip
2. Let $\Pi_2$ be the program consisting of the following rules:
\begin{eqnarray}
\label{eqn:pi2_1}\!\!\!\!\!\!\!\!y=c_{\epsilon}\!\!\!\!&\rightarrow&\!\!\!\!\textsc{mrg}(x,y,x)\\
\label{eqn:pi2_2}\!\!\!\!\!\!\!\!\left[\begin{aligned}
\textsc{mrg}(x,u,v)&\wedge\textsc{enc}(u,w,y)\\
&\wedge\textsc{enc}(v,w,z)
\end{aligned}\right]
\!\!\!\!\!\!&\rightarrow&\!\!\!\!\textsc{mrg}(x,y,z)\\
\label{eqn:pi2_3}\!\!\!\!\!\!\!\!\textsc{enc}(x,u,y)\!\!\!\!&\rightarrow&\!\!\!\!\textsc{in}(u,y)\\
\label{eqn:pi2_4}\!\!\!\!\!\!\!\!\textsc{enc}(x,z,y)\wedge\textsc{in}(u,x)\!\!\!\!&\rightarrow&\!\!\!\!\textsc{in}(u,y)\\
\label{eqn:pi2_5}\!\!\!\!\!\!\!\!x=c_{\epsilon}\!\!\!\!&\rightarrow&\!\!\!\!\textsc{subc}(x,y)\\
\label{eqn:pi2_6}\!\!\!\!\!\!\!\!\!\!\textsc{subc}(u,y)\wedge\textsc{enc}(u,v,x)\wedge\textsc{in}(v,y)\!\!\!\!&\rightarrow&\!\!\!\!\textsc{subc}(x,y)\\
\label{eqn:pi2_7}\!\!\!\!\!\!\!\!\!\!\textsc{subc}(x,y)\wedge\textsc{subc}(y,x)\!\!\!\!&\rightarrow&\!\!\!\!\textsc{equ}(x,y)
\end{eqnarray}
Informally, rules (\ref{eqn:pi2_1})--(\ref{eqn:pi2_2}) describe that $\textsc{mrg}$ is the graph of the merging function related to $\textsc{enc}$ and $c_{\epsilon}$; rules (\ref{eqn:pi2_3})--(\ref{eqn:pi2_4}) are an inductive version of  (\ref{eqn:defn_in}); rules (\ref{eqn:pi2_5})--(\ref{eqn:pi2_6}) are an inductive version of (\ref{eqn:defn_subc}); all rules (\ref{eqn:pi2_3})--(\ref{eqn:pi2_7}) then assert that $\textsc{equ}$ is the equivalence between positive clauses.

\smallskip
3. Let $\Pi_3$ be the logic program consisting of the rule
\begin{eqnarray}
\label{eqn:pi3_equtran}
\textsc{true}(x)\wedge\textsc{equ}(x,y)&\rightarrow&\textsc{true}(y)
\end{eqnarray}
and the rule
{\small
\begin{eqnarray}
\label{eqn:pi3_progsim}
\!\!\!\!\!\!\!\!\!\!\!\left[\begin{aligned}
&\textsc{true}(x_1)\wedge\cdots\wedge\textsc{true}(x_k)\,\wedge\\
&\textsc{enc}(y_1,\lceil\vartheta_1\rceil,x_1)
\wedge\cdots\wedge\textsc{enc}(y_k,\lceil\vartheta_k\rceil,x_k)\\
&\qquad\qquad\,\,\,\,\wedge\textsc{mrg}(\lceil\gamma_{\textsc{h}}\rceil,y_1,\dots,y_k,z)\wedge\gamma^{\overline{\mbox{ }\!\mbox{ }}}_{\textsc{b}}\end{aligned}\right]
\!\!\!\!\!\!\!&\rightarrow&\!\!\!\!\!\textsc{true}(z)
\end{eqnarray}
}

\noindent
for each rule $\gamma$ in $\Pi$, where $\vartheta_1,\dots,\vartheta_k$ list all the intensional atoms that have strictly positive occurrences in the body of $\gamma$ for some $k\geq 0$; $\gamma_{\textsc{h}}$ is the head of $\gamma$, $\gamma^-_{\textsc{b}}$ is the conjunction of literals occurring in the body of $\gamma$ but not in $\vartheta_1,\dots,\vartheta_k$; $\textsc{enc}(t_1,\lceil\vartheta\rceil,t_2)$ denotes the following conjunction
\begin{align*}
u^i_0=c_P\,\wedge\,&\textsc{enc}(u^i_0,t_1,u^i_1)\wedge\cdots\\
&\wedge\textsc{enc}(u^i_{m-1},t_m,u^i_m)\wedge\textsc{enc}(t_1,u^i_m,t_2)
\end{align*}
for some new variables $u^i_j$ if $t_1,t_2$ are terms and $\vartheta$ an atom of form $P(t_1,\dots,t_m)$;  $\textsc{mrg}(\lceil\gamma_{\textsc{h}}\rceil,y_1,\dots,y_k,z)$ denotes
\begin{align*}
\textsc{enc}(v_0,&\lceil\zeta_1\rceil,v_1)\wedge\cdots\wedge\textsc{enc}(v_{n-1},\lceil\zeta_n\rceil,v_n)\\
&\wedge\textsc{mrg}(w_0,y_1,w_1)\wedge\cdots\wedge\textsc{mrg}(w_{k-1},y_k,w_k)\\
&\wedge v_0=c_{\epsilon}\wedge w_0=v_n\wedge z=w_k
\end{align*}
if $\gamma_{\textsc{h}}=\zeta_1\vee\cdots\vee\zeta_n$ for some atoms $\zeta_1,\dots,\zeta_n$ and $n\geq 0$.

Intuitively, rule (\ref{eqn:pi3_equtran}) assures that the progression is closed under the equivalence of positive clauses; rule (\ref{eqn:pi3_progsim}) then simulates the progression operator for the original program. As each positive clause is encoded by an element in the intended domain, the processes of decoding and encoding should be carried out before and after the simulation respectively.

\begin{exm}
Let $\gamma\!=\!P(v)\wedge\neg Q(v)\rightarrow R(v)\vee S(v)$ be a rule such that $P,Q,R,S$ are intensional. Then the following rule, defined by\! (\ref{eqn:pi3_progsim})\! with a slight simplification, simulates $\gamma$:
{\small
\begin{equation*}
\!\!\left[\begin{aligned}
\!&\textsc{true}(x_1)\wedge\textsc{enc}(c_P,\!v,\!u_1)\wedge\textsc{enc}(y_1,\!u_1,\!x_1)\,\wedge\\
\!&\textsc{enc}(c_R,\!v,\!u_2)\wedge\textsc{enc}(c_{\epsilon},\!u_2,\!w_1)\wedge\textsc{enc}(c_S,\!v,\!u_3)\,\wedge\\
&\quad\,\qquad\textsc{enc}(w_1,\!u_3,\!w_2)\wedge
\textsc{mrg}(w_2,\!y_1,\!z)\wedge\neg Q(v)
\end{aligned}\right]\!\!\rightarrow\!\textsc{true}(z).
\end{equation*}
}
\end{exm}

\smallskip
4. Let $\Pi_4$ be the program consisting of the rule
\begin{eqnarray}
\label{eqn:pi4_false1}x=c_{\epsilon}&\rightarrow&\textsc{false}(x)
\end{eqnarray}
and the rule
\begin{eqnarray}
\label{eqn:pi4_false2}\!\!\!\!\!\!\!\textsc{false}(x)\wedge\textsc{enc}(x,\lceil\vartheta\rceil,y)\wedge\neg\vartheta\!\!\! &\rightarrow&\!\!\! \textsc{false}(y)
\end{eqnarray}
for every intensional atom $\vartheta$ of the form $P(\bar{z}_P)$, where $\bar{z}_P$ denotes a tuple of distinct individual variables $z_1\cdots z_{k_P}$ that are different from $x$ and $y$, and $k_P$ is the arity of $P$.

This program is intended to define the predicate $\textsc{false}$ as follows: $\textsc{false}(a)$ holds in the intended structure if and only if $a$ encodes a positive clause that is false in the structure. 

\smallskip
5. Let $\Pi_5$ be the logic program consisting of the rule
\begin{eqnarray}
\label{eqn:pi5_true1}\textsc{true}(c_{\epsilon})&\rightarrow&\bot
\end{eqnarray}
and the following rule
\begin{eqnarray}
\label{eqn:pi5_true2}\!\!\!\!\!\!\!\textsc{true}(x)\wedge\textsc{enc}(y,\lceil\vartheta\rceil,x)\wedge\textsc{false}(y)\!\!\! &\rightarrow&\!\!\!\vartheta
\end{eqnarray}
for each atom $\vartheta$ of the form same as that in $\Pi_4$.

Informally, this program asserts that a ground atom is true in the intended structure if and only if there is a positive clause containing this atom such that the clause in true and all the other atoms in this clause are false in the structure.

\smallskip
Now, we let $\Pi^{\diamond}$ denote the union of $\Pi_1,\dots,\Pi_5$. This then completes the definition of the translation. The soundness of this translation is assured by the following theorem.


\begin{thm}\label{thm:tran_infinite}
Let $\Pi$ be a disjunctive program. Then over infinite structures, $\textsc{SM}(\Pi)$ is equivalent to $\exists\pi\textsc{SM}(\Pi^{\diamond})$, where $\pi$ denotes the set of constants occurring in $\Pi^{\diamond}$ but not in $\Pi$.
\end{thm}

To prove this result, some notations and lemmas are needed.
Let $\upsilon_i$ and $\tau$ be the sets of intensional predicates of $\Pi_i$ and $\Pi$ respectively. Let $\sigma=\upsilon_1\cup\upsilon_2\cup\upsilon(\Pi)$. Given a structure $\mathds{A}$ of $\upsilon(\Pi)$, each {\em encoding expansion} of $\mathds{A}$ is defined to be a $\sigma$-expansion $\mathds{B}$ of $\mathds{A}$ satisfying both of the following:
\begin{enumerate}
\item $\textsc{enc}$ is interpreted as the graph of an encoding function $\textsl{enc}$ on $A$ such that no element among $c${\small$_{\epsilon}^{\mathds{B}}$} and $c${\small$_P^{\mathds{B}}$} (for all $P\in\tau$) belongs to the range of $\textsl{enc}$, $\underline{\textsc{enc}}$ as the complement of the graph of $\textsl{enc}$, and $\textsc{ok}${\small$_e^{\mathds{B}}$}$=A\times A$;
\item $\textsc{mrg}$ is interpreted as the graph of the merging function related to $\textsl{enc}$ and $c${\small$_{\epsilon}^{\mathds{B}}$}, and $\textsc{in},\textsc{subc},\textsc{equ}$ as the encoding predicates $\textsl{in},\textsl{subc},\textsl{equ}$ related to $\textsl{enc}$ and $c${\small$_{\epsilon}^{\mathds{B}}$} respectively.
\end{enumerate}

%

Let $\mathds{A}$ be a structure of $\upsilon(\Pi)$ with an encoding expansion $\mathds{B}$. Let $\textsl{enc}$ be the encoding function with graph $\textsc{enc}${\small$^{\mathds{B}}$}. Let
\begin{align}
[\![P(a_1\hspace{-0.005in},\hspace{-0.005in}\dots\hspace{-0.005in},\hspace{-0.005in}a_k)]\!]^{\mathds{B}}&=\textsl{enc}(a_1,\dots,a_k;c_P^{\mathds{B}}),\\
[\![\vartheta_1\vee\hspace{0.004in}\cdots\hspace{0.005in}\vee\vartheta_n]\!]^{\mathds{B}}&=\textsl{enc}([\![\vartheta_1]\!]^{\mathds{B}},\dots,[\![\vartheta_n]\!]^{\mathds{B}};c_{\epsilon}^{\mathds{B}}).
\end{align}
If the encoding expansion $\mathds{B}$ is clear from the context, we simply write $[\![\cdot]\!]${\small$^{\mathds{B}}$} as $[\![\cdot]\!]$. Given a set $\Sigma$ of ground positive clauses, let $[\![\Sigma]\!]$ be the set of elements $[\![C]\!]$ for all $C\in\Sigma$. Let $\Delta^n(\mathds{B})$ be the set of elements $a$ with $\textsc{true}(a)\in\Gamma${\small$^{\mathds{B}}_{\Pi_3}$}$\uparrow_n$.


\begin{lem}\label{lem:progress_sim}
Let $\mathds{A}$ be a structure of $\upsilon(\Pi)$ with an encoding expansion $\mathds{B}$. Then $[\![\Gamma${\small$^{\mathds{A}}_{\Pi}$}$\uparrow_{\omega}]\!]=\cup_{n\ge 0}\Delta${\small$^{n}$}$(\mathds{B})$.
\end{lem}

Next, given a structure $\mathds{A}$ of $\upsilon(\Pi)$, an $\upsilon(\Pi^{\diamond})$-expansion $\mathds{C}$ of $\mathds{A}$ is called a {\em progression expansion} of $\mathds{A}$ if the restriction of $\mathds{C}$ to $\sigma$, denoted by $\mathds{B}$, is an encoding expansion of $\mathds{A}$; $\mathds{C}$ interprets $\textsc{true}$ as $[\![\Gamma${\small$^{\mathds{A}}_{\Pi}$}$\uparrow_{\omega}]\!]${\small$^{\mathds{B}}$}, and interprets $\textsc{false}$ as
\begin{equation}
\left\{[\![C]\!]^{\mathds{B}}\mid C\in\textsc{gpc}(\tau,A)\,\,\&\,\,\textsc{Ins}(\mathds{A},\tau)\models\neg C\right\}
\end{equation}
where $\textsc{gpc}(\tau,A)$ denotes the set of ground positive clauses built from predicates in $\tau$ and elements in $A$.

\begin{lem}\label{lem:infinite_pro} Let $\mathds{A}$ be a structure of $\upsilon(\Pi)$ with a progression expansion $\mathds{C}$. Then $\textsc{Ins}(\mathds{A},\tau)$ is a minimal model of $\Gamma${\small$^{\mathds{A}}_{\Pi}$}$\uparrow_{\omega}$ iff $\textsc{Ins}(\mathds{C},\tau)$ is a minimal model of $\Pi${\small$_5^{\mathds{C}}$}.

\end{lem}

Due to the limit of space, we will omit the proofs of Lemmas \ref{lem:progress_sim} and \ref{lem:infinite_pro} here.
To show Lemma \ref{lem:progress_sim}, it is sufficient to show both $[\![\Gamma${\small$^{\mathds{A}}_{\Pi}$}$\uparrow_{n}]\!]\subseteq\Delta${\small$^{2n}$}$(\mathds{B})$ and $[\![\Gamma${\small$^{\mathds{A}}_{\Pi}$}$\uparrow_{n}]\!]\supseteq\Delta${\small$^{n}$}$(\mathds{B})$, and each of them can be done by an induction. For Lemma \ref{lem:infinite_pro}, roughly speaking, the soundness is assured by the result that every head-cycle-free disjunctive program is equivalent to the normal program obtained by shifting~\cite{BD:94}. Please note that every set of positive clauses is head-cycle-free, and $\Pi_4$ and $\Pi_5$ are designed for the simulation of shifting.
With these lemmas, we can then prove Theorem \ref{thm:tran_infinite}:

\begin{proof}[Proof of Theorem \ref{thm:tran_infinite}]
By the splitting lemma in~\cite{FLLP:09} and the second-order transformation, it suffices to show that $\mathrm{SM}(\Pi)$ is equivalent to the following formula
\begin{equation}\label{eqn:splitting}
\exists\pi[\mathrm{SM}(\Pi_1)\wedge\cdots\wedge\mathrm{SM}(\Pi_5)]
\end{equation}
over infinite structures. Now we prove it as follows.

``$\Longrightarrow$": Let $\mathds{A}$ be an infinite model of $\mathrm{SM}(\Pi)$. Let $\mathds{B}$ be an encoding expansion of $\mathds{A}$. The existence of such an expansion is clearly assured by the infiniteness of $A$. It is easy to check that $\mathds{B}$ is a stable model of both $\Pi_1$ and $\Pi_2$. Let $\mathds{C}$ be the progression expansion of $\mathds{A}$ that is also an expansion of $\mathds{B}$. By Proposition \ref{prop:fxp2sm}, $\textsc{Ins}(\mathds{B},\tau)=\textsc{Ins}(\mathds{A},\tau)$ should be a minimal model of $\Gamma${\small$^{\mathds{A}}_{\Pi}$}$\uparrow_{\omega}$. By Lemma \ref{lem:progress_sim} and definition, $\textsc{Ins}(\mathds{B},\tau)$ is also a minimal model of $\Gamma${\small$^{\mathds{B}}_{\Pi_3}$}$\uparrow_{\omega}$. By Proposition \ref{prop:fxp2sm} again, $\mathds{B}$ is then a stable model of $\Pi_3$, which implies that so is $\mathds{C}$. It is also easy to check that $\mathds{C}$ is a stable model of $\Pi_4$. On the other hand, since $\textsc{Ins}(\mathds{A},\tau)$ is a minimal model of $\Gamma${\small$^{\mathds{A}}_{\Pi}$}\,$\uparrow_{\omega}$, by Lemma \ref{lem:infinite_pro}, $\textsc{Ins}(\mathds{C},\tau)$ should be a minimal model of $\Pi${\small$_5^{\mathds{C}}$}, which means that $\mathds{C}$ is a stable model of $\Pi_5$ by Proposition \ref{prop:fo2prop}. Thus, $\mathds{A}$ is a model of formula (\ref{eqn:splitting}).

``$\Longleftarrow$": Let $\mathds{A}$ be an infinite model of formula (\ref{eqn:splitting}). Then there exists an $\upsilon(\Pi^{\diamond})$-expansion $\mathds{C}$ of $\mathds{A}$ such that $\mathds{C}$ satisfies $\mathrm{SM}(\Pi_i)$ for all $i:1\le i\le 5$. Let $\mathds{B}$ be the restrictions of $\mathds{C}$ to $\sigma$. Then, by a routine check, it is easy to show that $\mathds{B}$ is an encoding expansion of $\mathds{A}$. As $\mathds{C}$ is a stable model of $\Pi_3$, by Proposition \ref{prop:fxp2sm}, $\textsc{Ins}(\mathds{C},\upsilon_3)$ is then a minimal model of $\Gamma${\small$^{\mathds{C}}_{\Pi_3}$}\,$\uparrow_{\omega}\,=\Gamma${\small$^{\mathds{B}}_{\Pi_3}$}\,$\uparrow_{\omega}$. Furthermore, by Lemma \ref{lem:progress_sim} and the conclusion that $\mathds{C}$ satisfies $\mathrm{SM}(\Pi_4)$, we then have that $\mathds{C}$ is a progression expansion of $\mathds{A}$. On the other hand, since $\mathds{C}$ is also a stable model of $\Pi_5$, by Proposition \ref{prop:fo2prop} we can conclude that $\textsc{Ins}(\mathds{C},\tau)$ is a minimal model of $\Pi${\small$_5^{\mathds{C}}$}. Thus, by Lemma \ref{lem:infinite_pro} we immediately have that $\textsc{Ins}(\mathds{A},\tau)$ is a minimal model of $\Gamma${\small$^{\mathds{A}}_{\Pi}$}\,$\uparrow_{\omega}$. By Proposition \ref{prop:fxp2sm}, $\mathds{A}$ is then a stable model of $\Pi$.
%
%
\end{proof}

\begin{rem}
Note that, given any finite domain $A$, there is no injective function from $A\times A$ into $A$. Therefore, we can not expect that the above translation works on finite structures.
\end{rem}


\begin{cor}
$\mathrm{DLP}\simeq_{\mathsf{INF}}\mathrm{NLP}$.
\end{cor}


Now, let us focus on the relationship between logic programs and second-order logic. The following proposition says that,  over infinite structures, normal programs are more expressive than the existential second-order logic, which then strengthens a result in~\cite{ALZZ:12} where such a separation over arbitrary structures was obtained.

\begin{prop}\label{prop:infinite_nlp2eso}
$\mathrm{NLP}\not\le_{\mathsf{INF}}\Sigma^1_1$.
\end{prop}

To show this, our main idea is to define a property that can be defined by a normal program but not by any existential second-order sentence. The property is defined as follows. Let $\upsilon_{\textsc{r}}$ be the vocabulary consisting of a binary predicate $E$ and two individual constants $s$ and $t$. Let $\textsc{Reach}_{i}$ be the class of infinite $\upsilon_{\textsc{r}}$-structures in each of which there is a finite path from $s$ to $t$ via edges in $E$. Now, we show the result.

\begin{proof}[Proof of Proposition \ref{prop:infinite_nlp2eso}]
First show that $\textsc{Reach}_{i}$ is definable in $\mathrm{NLP}$ over infinite structures.
Let $\Pi$ be the normal program
\begin{equation}
\left\{\,P(s)^{^{}}_{_{}},\,\, P(x)\wedge E(x,y)\rightarrow P(y),\,\, \neg P(t)\rightarrow\bot\,\right\}.
\end{equation}
By a simple check, we can show that the formula $\exists P\mathrm{SM}(\Pi)$ defines the desired property over infinite structures.

Next, we prove that $\textsc{Reach}_{i}$ is undefinable in $\Sigma${\small$^1_1$} over infinite structures. Towards a contradiction, assume that there is a first-order sentence $\varphi$ and a finite set $\tau$ of predicates such that $\exists\tau\varphi$ is in $\Sigma${\small$^1_1$} and defines $\textsc{Reach}_{i}$ over infinite structures. Let $R$ be a binary predicate not in $\tau$. Let $\psi$ denote
\begin{equation}
\begin{aligned}
\forall x\exists y R(x&,y)\wedge\forall x\neg R(x,x)\wedge\\
\forall &x\forall y\forall z[R(x,y)\wedge R(y,z)\rightarrow R(x,z)].
\end{aligned}
\end{equation}
Intuitively, it asserts that the relation $R$ is both transitive and irreflexive, and each element in the domain has a successor w.r.t. this relation. It is obvious that such a relation exists if and only if the domain is infinite. Therefore, the formula $\exists\tau\varphi\wedge\exists R\psi$ defines $\textsc{Reach}_i$ over arbitrary structures.

Moreover, let $\gamma_0(x,y)$ be $x=y$; for all $n>0$ let $\gamma_n(x,y)$ denote $\exists z_n(\gamma_{n-1}(x,z_n)\wedge E(z_n,y))$, where each $\gamma_n(x,y)$ asserts that there is a path of length $n$ from $x$ to $y$. Let $\Lambda$ be the set of sentences $\neg\gamma_n(s,t)$ for all $n\ge 0$. Now we claim:

\smallskip
\noindent{\em Claim.} $\Lambda\cup\{\exists\tau\varphi,\exists R\psi\}$ is satisfiable.
\smallskip

To show this, it suffices to show that the first-order theory $\Lambda\cup\{\varphi,\psi\}$ is satisfiable. Let $\Phi$ be a finite subset of $\Lambda$, and let $n=\max\{m\mid\neg\gamma_m(s,t)\in\Phi\}$. Let $\mathds{A}$ be an infinite model of $\psi$ with vocabulary $\upsilon(\varphi)\cup\upsilon(\psi)$ in which the minimal length of paths from $s$ to $t$ via edge $E$ is an integer $>n$. Then $\mathds{A}$ is clearly a model of $\Phi\cup\{\varphi,\psi\}$. Due to the arbitrariness of $\Phi$, by the compactness we then have the desired claim.

\smallskip
Let $\mathds{A}$ be any model of $\Lambda\cup\{\exists\tau\varphi,\exists R\psi\}$. Then according to $\exists R\psi$, $\mathds{A}$ should be infinite, and by $\Lambda$, there is no path from $s$ to $t$ via $E$ in $\mathds{A}$. However, according to $\exists\tau\varphi$, every infinite model of it should be $s$-to-$t$ reachable, a contradiction. Thus, the property $\textsc{Reach}_i$ is then undefinable in $\Sigma^1_1$ over infinite structures. This completes the proof immediately.
%
\end{proof}

The following separation immediately follows from the proof of Theorem 4.1 in~\cite{EGG:96}. Although their statement refers to arbitrary structures, the proof still works if only infinite structures are focused.

\begin{prop}
$\Sigma^{1}_{2}\not\le_{\mathsf{INF}}\mathrm{DLP}$.
\end{prop}

\section{Finite Structures}

This section will focus on the expressiveness of logic programs over finite structures. We first consider 
%
%
the relationship between disjunctive and normal programs. Unfortunately, in the general case, we have the following result:

\begin{prop}\label{prop:fin_dlp2nlp}
$\mathrm{DLP}\simeq_{\mathsf{FIN}}\mathrm{NLP}$ iff $\mathrm{NP}=\mathrm{coNP}$.\footnote{A similar result for traditional logic programs under the query equivalence can be obtained by the expressiveness results proved by~\cite{Schl:95,EGM:97}.}
\end{prop}

\begin{proof}
By Fagin's Theorem~\cite{Fagin:74} and Stockmeyer's logical characterization of the polynomial hierarchy~\cite{Stoc:77},\footnote{In their characterizations of complexity classes, no function constant of positive arity is allowed. However, this restriction can be removed as functions can be easily simulated by predicates.} we have that $\Sigma^1_2\simeq_{\mathsf{FIN}}\Sigma^1_1$ iff $\Sigma^p_2=\mathrm{NP}$. By a routine complexity theoretical argument, it is also true that $\Sigma^p_2=\mathrm{NP}$ iff $\mathrm{NP}=\mathrm{coNP}$. On the other hand, according to the proof of Theorem 6.3 in~\cite{EGM:97}, or by Proposition \ref{prop:so2dlp} in this section, Leivant's normal form~\cite{Leiv:89} and the definition of $\mathrm{SM}$, we can conclude $\mathrm{DLP}\simeq_{\mathsf{FIN}}\Sigma^1_2$;
by Proposition \ref{prop:nlp2eso} in this section, it holds that $\mathrm{NLP}\simeq_{\mathsf{FIN}}\Sigma^1_1$. Combining these conclusions, we then have the desired proposition.
\end{proof}

This result shows us how difficult it is to separate normal programs from disjunctive programs over finite structures. To know more about the relationship, we will try to prove a weaker separation between these two classes. Before doing this, we need to study the relationship between logic programs and second-order logic. For the class of normal programs, we have the following characterization:


\begin{prop}\label{prop:nlp2eso}
$\mathrm{NLP}_n^{\textsc{f}}\simeq_{\mathsf{FIN}}\Sigma_{1,n}^{1\textsc{f}}[\forall^{\ast}]$ for all $n>1$.
\end{prop}

%

To prove the above characterization, we have to develop a translation that turns normal programs to first-order sentences. The main idea is to extend the Clark completion by a progression simulation, so it is an improved version of the ordered completion proposed by~\cite{ALZZ:12}. 

Now, we define the translation. Let $\Pi$ be a normal program and $n$ the maximal arity of intensional predicates of $\Pi$. Without loss of generality, assume the head of every rule in $\Pi$ is of form $P(\bar{x})$, where $P$ is a $k$-ary predicate for some $k\ge 0$, and $\bar{x}$ is the tuple of distinct individual variables $x_1,\dots,x_k$. Let $\prec$ be a new binary predicate and $\varpi$ a universal first-order sentence asserting that $\prec$ is a strict partial order. Given two tuple $\bar{s},\bar{t}$ of terms of the same length, let $\bar{s}\prec\bar{t}$ be a quantifier-free formula asserting that $\bar{s}$ is less than $\bar{t}$ w.r.t. the lexicographic order extended from $\prec$ naturally.


Let $\tau$ be the set of intensional predicates of $\Pi$. Let $c$ be the least integer $\ge\log_2|\tau|+n$. Fix $P$ to be a $k$-ary predicate in $\tau$ and let $\lambda=P(x_1,\dots,x_k)$. Suppose $\gamma_1,\dots,\gamma_{l}$ list all the rules in $\Pi$ whose heads are $\lambda$, and suppose $\gamma_i$ is of form
\begin{equation}\label{eqn:nlp2eso_rule}
\zeta^i\wedge\vartheta^i_1\wedge\cdots\wedge\vartheta^i_{m_i}\rightarrow\lambda
\end{equation}
where {\small$\vartheta^i_1,\dots,\vartheta^i_{m_i}$} list all the positive intensional conjuncts in the body of $\gamma_i$, {\small$\zeta^i$} is the conjunction of other conjuncts that occurs in the body of $\gamma_i$, $m_i\geq 0$, and $\bar{y}_i$ is the tuple of all individual variables occurring in $\gamma_i$ but not in $\lambda$.

Next, we let $\varphi_{P}$ denote the conjunction of rules $\gamma_i$
for all $i:1\leq i\leq l$, and let $\psi_{P}$ denote the formula
\begin{equation}\label{eqn:nlp2eso_completion2}
\lambda\rightarrow\bigvee_{i=1}^{l}\exists\bar{y}_i\left[\zeta^i\wedge\bigwedge_{j=1}^{m_i}\left(\vartheta^i_j\wedge \textsc{less}_{\textsc{d}}(\vartheta^i_j,\lambda)\right)\right]
\end{equation}
where, for every intensional atoms $\vartheta$ and $\vartheta_0$, $\mathrm{ord}(\vartheta)$ denotes the tuple $(o${\small$_Q^c$}$(\bar{t}),\cdots,o${\small$_Q^{1}$}$(\bar{t}))$ if $\vartheta$ of form $Q(\bar{t})$, each $o${\small$_Q^s$} is a new function whose arity is the same as that of $Q$, and $\textsc{less}_{\textsc{d}}(\vartheta,\vartheta_0)$ denotes formula $\mathrm{ord}(\vartheta)\prec\mathrm{ord}(\vartheta_0)$.

Define $\varphi_{\Pi}$ as the universal closure of conjunction of the formula $\varpi$ and formulas $\varphi_{P}\wedge\psi_{P}$ for all $P\in\tau$. Let $\sigma$ be the set of functions $o${\small$^s_{Q}$} for all $Q\in\tau$ and $s\hspace{-.02cm}:1\hspace{-.02cm}\le\hspace{-.02cm}s\hspace{-.02cm}\le\hspace{-.02cm}c$. Clearly, $\exists\sigma\varphi_{\Pi}$ is equivalent to a sentence in $\Sigma${\small$^{1{\textsc{f}}}_{1,n}$}$[\forall^{\ast}]$ by introducing Skolem functions if $n>1$. Now we show the soundness: 

\begin{lem}\label{lem:finite_nlp2eso}
Given any finite structure $\mathds{A}$ of $\upsilon(\Pi)$ with at least two elements in the domain, $\mathds{A}\models\mathrm{SM}(\Pi)$ iff $\mathds{A}\models\exists\sigma\varphi_{\Pi}$.
\end{lem}

\begin{proof}
%
Due to the limit of space, we only show the right-to-left direction. Let $\mathds{B}$ be a finite model of $\varphi_{\Pi}$. By formula $\varpi$, $\mathds{B}$ must interpret predicate $\prec$ as a strict partial order on $B$. Let $\mathds{A}$ be the restriction of $\mathds{B}$ to $\upsilon(\Pi)$. To show that $\mathds{A}$ is a stable model of $\Pi$, by Proposition \ref{prop:fxp2sm} it suffices to show that $\textsc{Ins}(\mathds{A},\tau)=\Gamma${\small$^{\mathds{A}}_{\Pi}$}\,$\uparrow_{\omega}$. We first claim that $\Gamma${\small$^{\mathds{A}}_{\Pi}$}\,$\uparrow_{m}\,\subseteq\textsc{Ins}(\mathds{A},\tau)$ for all $m\geq 0$. This can be shown by an induction on $m$. The case of $m=0$ is trivial. Let $m>0$ and assume $\Gamma${\small$^{\mathds{A}}_{\Pi}$}\,$\uparrow_{m-1}\,\subseteq\textsc{Ins}(\mathds{A},\tau)$. Our task is to show $\Gamma${\small$^{\mathds{A}}_{\Pi}$}\,$\uparrow_{m}\,\subseteq\textsc{Ins}(\mathds{A},\tau)$. Let $p$ be a ground atom in $\Gamma${\small$^{\mathds{A}}_{\Pi}$}\,$\uparrow_{m}$. By definition, there must exist a rule $\gamma_i$ of form (\ref{eqn:nlp2eso_rule}) in $\Pi$ and an assignment $\alpha$ in $\mathds{A}$ such that $\lambda[\alpha]=p$, $\alpha$ satisfies $\zeta^i$ in $\mathds{A}$ (so equivalently, in $\mathds{B}$), and for each atom $\vartheta${\small$^i_j$}, $\vartheta${\small$^i_j$}$[\alpha]\in\Gamma${\small$^{\mathds{A}}_{\Pi}$}\,$\uparrow_{m-1}$. By the inductive assumption, each $\vartheta${\small$^i_j$}$[\alpha]\in\textsc{Ins}(\mathds{A},\tau)$, or in other words, $\alpha$ satisfies each $\vartheta${\small$^i_j$} in $\mathds{A}$ (so equivalently, in $\mathds{B}$). As $\alpha$ clearly satisfies the rule $\gamma_i$ in $\mathds{B}$, we can conclude that $\alpha$ satisfies $\lambda$ in $\mathds{B}$, which implies  $p=\lambda[\alpha]\in\textsc{Ins}(\mathds{B},\tau)=\textsc{Ins}(\mathds{A},\tau)$. So, the claim is true. From it, we have $\Gamma^{\mathds{A}}_{\Pi}\uparrow_{\omega}\subseteq\textsc{Ins}(\mathds{A},\tau)$.

Now, it remains to prove $\textsc{Ins}(\mathds{A},\tau)\subseteq\Gamma${\small$^{\mathds{A}}_{\Pi}$}\,$\uparrow_{\omega}$.
Towards a contradiction, assume this is not true. Then we must have $\Gamma${\small$^{\mathds{A}}_{\Pi}$}\,$\uparrow_{\omega}\,\subsetneq\textsc{Ins}(\mathds{A},\tau)$ by the previous conclusion.
Given two ground intensional atoms $p_1$ and $p_2$ in $\textsc{Ins}(\mathds{A},\tau)$, we define $p_1<p_2$ if $\textsc{less}_{\textsc{d}}(p_1,p_2)$ is true in $\mathds{B}$. Let $p$ be a $<$-minimal atom in $\textsc{Ins}(\mathds{A},\tau)-\Gamma${\small$^{\mathds{A}}_{\Pi}$}\,$\uparrow_{\omega}$ and suppose $p=P(\bar{a})$ for some $P\in\tau$. Let $\alpha$ be an assignment in $\mathds{B}$ such that $\alpha(\bar{x})=\bar{a}$. By definition, $\alpha$ should satisfy $\psi_{P}$ (in which $\lambda[\alpha]=p$) in $\mathds{B}$. So, there exist an integer $i:1\le i\le l$ and an assignment $\alpha_0$ in $\mathds{B}$ such that (i) $\alpha_0(\bar{x})=\bar{a}$, (ii) $\zeta${\small$^i$}$[\alpha_0]$ is true in $\mathds{B}$, and (iii) for all $j$, $q_j\in\textsc{Ins}(\mathds{B},\tau)=\textsc{Ins}(\mathds{A},\tau)$ and $q_j<\lambda[\alpha_0]$, where $q_j$ denotes $\vartheta${\small$^i_j$}$[\alpha_0]$. As $\lambda[\alpha_0]=\lambda[\alpha]=p$ and $p$ is $<$-minimal  in $\textsc{Ins}(\mathds{A},\tau)-\Gamma${\small$^{\mathds{A}}_{\Pi}$}\,$\uparrow_{\omega}$, we can conclude $q_j\in\Gamma${\small$^{\mathds{A}}_{\Pi}$}\,$\uparrow_{\omega}$ for all $j$. According to the definition of $\psi_P$, the rule $\gamma_i$ (of form (\ref{eqn:nlp2eso_rule})) is in $\Pi$, which implies $q_1\wedge\cdots\wedge q_{m_i}\rightarrow p=\gamma_i^+[\alpha_0]\in\Pi^{\mathds{A}}$. By definition, we then have $p\in\Gamma${\small$^{\mathds{A}}_{\Pi}$}\,$\uparrow_{\omega}$, a contradiction. 
\end{proof}


\begin{rem}
Let $m$ and $n$ be the number and the maximal arity of intensional predicates respectively. The maximal arity of auxiliary constants in our translation is only $n$ (that of the ordered completion in~\cite{ALZZ:12} is $2n$), which is optimal if Conjecture 1\footnote{It implies $\mathrm{ESO}${\small$^{\textsc{f}}_n$}$[\forall^{\ast}]\simeq_{\mathsf{FIN}}\mathrm{ESO}${\small$^{\textsc{f}}_n$}$[\forall^{n}]$, where the latter is the class of sentences in $\mathrm{ESO}${\small$^{\textsc{f}}_n$}$[\forall^{\ast}]$ with at most $n$ individual variables.} in~\cite{DGO:04} is true. Moreover, the number of auxiliary constants in our translation is $m\cdot(\lceil\log_2 m\rceil+n)$, while that of the ordered completion is $m^2$. (Note that $n$ is normally very small.) 
\end{rem}

\begin{rem}
Similar to the work in~\cite{ALZZ:12}, we can develop an answer set solver by calling some SMT solver. By the comparison in the above remark, this approach is rather promising. In addition, as a strict partial order is available in almost all the SMT solvers (e.g., built-in arithmetic relations), our translation can be easily optimized.
\end{rem}

Now we are in the position to prove Proposition \ref{prop:nlp2eso}.

\begin{proof}[Proof of Proposition \ref{prop:nlp2eso}]
``$\geq_{\mathsf{FIN}}$": Let $\varphi$ be any sentence in $\Sigma${\small$^{1{\textsc{f}}}_{1,n}$}$[\forall^{\ast}]$. It is obvious that $\varphi$ can be written as an equivalent sentence of form $\exists\tau\forall\bar{x}(\gamma_1\wedge\cdots\wedge\gamma_k)$ for some $k\geq 0$, where each $\gamma_i$ is a disjunction of atoms or negated atoms, and $\tau$ a finite set of functions or predicates of arity $\leq n$. Let $\Pi$ be a logic program consisting of the rule $\tilde{\gamma}_i\rightarrow\bot$ for each $i:1\leq i\leq k$, where $\tilde{\gamma}_i$ is obtained from $\gamma_i$ by substituting $\vartheta$ for each negated atom $\neg\vartheta$, followed by substituting $\neg\vartheta$ for each atom $\vartheta$, and followed by substituting $\wedge$ for $\vee$. It is easy to check that $\exists\tau\mathrm{SM}(\Pi)$ is in $\mathrm{NLP}${\small$_n^{\textsc{f}}$} and equivalent to $\varphi$.

``$\leq_{\mathsf{FIN}}$": Let $\mathcal{C}${\small$^{=1}$} (respectively, $\mathcal{C}${\small$^{>1}$}) be the class of finite structures with exactly one (respectively, at least two) element(s) in the domain. Let $\Pi$ be a normal program and $\tau$ a finite set of predicates and functions such that $\exists\tau\mathrm{SM}(\Pi)$ is in $\mathrm{NLP}${\small$_n^{\textsc{f}}$}. It is trivial to construct a sentence, say $\zeta$, in $\Sigma${\small$^{1{\textsc{f}}}_{1,n}$}$[\forall^{\ast}]$ such that $\exists\tau\mathrm{SM}(\Pi)$ is equivalent to $\zeta$ over $\mathcal{C}^{=1}$. (Please note that, if the domain is a singleton, a first-order logic program will regress to a propositional one.) By Lemma \ref{lem:finite_nlp2eso}, there is also a sentence $\psi$ in $\Sigma${\small$^{1{\textsc{f}}}_{1,n}$}$[\forall^{\ast}]$ such that $\exists\tau\mathrm{SM}(\Pi)$ is equivalent to $\psi$ over $\mathcal{C}${\small$^{>1}$}. Let $\varphi$ be the following sentence:
\begin{equation}
[\exists x\forall y(x=y)\wedge\zeta]\vee[\exists x\exists z(\neg x=z)\wedge\psi].
\end{equation}
Informally, this formula first test whether or not the domain is a singleton. If it is true, let $\zeta$ work; otherwise let $\psi$ work. Thus, it is easy to show that $\exists\tau\mathrm{SM}(\Pi)$ is equivalent to $\varphi$ over finite structures. It is also clear that $\varphi$ can be written to be an equivalent sentence in $\Sigma${\small$^{1{\textsc{f}}}_{1,n}$}$[\forall^{\ast}]$. (Please note that every first-order quantifier can be regarded as a second-order quantifier over a function variable of arity $0$.)
%
%
%
\end{proof}

\begin{rem}
Assuming Conjecture 1 in~\cite{DGO:04}, by the results of~\cite{Gran85}, $\mathrm{NLP}_k^{\textsc{f}}$ then exactly captures the class of languages computable in $O(n^{k})$-time (where $n$ is the size of input) in Nondeterministic Random Access Machines (NRAMs), and whether an extensional database can be expanded to a stable model of a disjunctive program is decidable in $O(n^k)$-time in NRAMs.
\end{rem}


By Proposition \ref{prop:nlp2eso} and the fact that auxiliary functions can be simulated by auxiliary predicates in both logic programs and second-order logic, we have the following result:

\begin{cor}
$\mathrm{NLP}\simeq_{\mathsf{FIN}}\Sigma^1_1$.
\end{cor}

\smallskip
Next, let us focus on the translatability from a fragment of second-order logic to disjunctive programs.
For convenience, in the rest of this paper, we fix $\textsc{succ}$ to be a binary predicate, fix $\textsc{first}$ and $\textsc{last}$ to be two unary predicates, and fix $\upsilon_{\textsc{s}}$ to be the set consisting of these predicates. In particular, unless mentioned otherwise, a logic program or a formula is always assumed to contains no predicate in $\upsilon_{\textsc{s}}$.

\smallskip
%
A structure $\mathds{A}$ is called a {\em successor structure} if:
\begin{enumerate}
\item its vocabulary contains all the predicates in $\upsilon_{\textsc{s}}$, and
\item $\textsc{succ}^{\mathds{A}}$ is a binary relation $R$ on $A$ such that the transitive closure of $R$ is a strict total order and for all $a\in A$, both $|\{b\!\mid\!(a,b)\in R\}|\le 1$ and $|\{b\!\mid\!(b,a)\in R\}|\le 1$ hold, and
\item $\textsc{first}^{\mathds{A}}$ (respectively, $\textsc{last}^{\mathds{A}}$) consists of the least element (respectively, the largest element) in $A$ w.r.t. $\textsc{succ}^{\mathds{A}}$.
\end{enumerate}
By this definition, given a successor structure, both the least and largest elements must exist, so it is then finite. Now, we let {\small$\textsf{SUC}$} denote the class of successor structures.
\smallskip

Let $\Sigma${\small$^1_{2,n}$}\hspace{-.02cm}$[\forall^n\exists^{\ast}]$ be the class of sentences in $\Sigma${\small$^1_{2,n}$}\hspace{-.02cm}$[\forall^{\ast}\exists^{\ast}]$ that involve at most $n$ universal quantifiers. Now we can show:

\begin{lem}\label{lem:so2dlp}
$\Sigma^{1}_{2,n}[\forall^n\exists^{\ast}]\leq_{\mathsf{SUC}}\mathrm{DLP}_n$ for all $n>0$.
\end{lem}

\begin{proof} (Sketch)
Let $\exists\tau\forall\sigma\varphi$ be any sentence in $\Sigma^{1}_{2,n}[\forall^n\exists^{\ast}]$ where $\tau,\sigma$ are finite sets of predicates of arities $\leq n$. Without loss of generality, suppose $\varphi=\forall\bar{x}\exists\bar{y}(\vartheta_1\vee\cdots\vee\vartheta_m)$, where $\bar{x}$ is of length $n$; each $\vartheta_i$ is a finite conjunction of literals. Next, we want to construct a disjunctive program which defines the property expressed by the sentence $\exists\tau\forall\sigma\varphi$.

Before constructing the program, we need to define some notations. Let $\bar{u}$ and $\bar{v}$ be any two tuples of individual variables $u_1\cdots u_k$ and $v_1\cdots v_k$ respectively. Let $\textsc{First}(\bar{u})$ denote the conjunction of $\textsc{first}(u_i)$ for all $i:1\le i\le k$, and let $\textsc{Last}(\bar{u})$ denote the conjunction of $\textsc{last}(u_i)$ for all $i:1\le i\le k$. Moreover, let $\textsc{Succ}_i(\bar{u},\bar{v})$ be the formula
\begin{eqnarray}
\left[
\begin{aligned}
&u_1=v_1\wedge\cdots\wedge u_{i-1}=v_{i-1}\\
&\quad\wedge\textsc{succ}(u_{i},v_{i})\wedge\textsc{last}(u_{i+1})\\
&\quad\wedge\textsc{first}(v_{i+1})\wedge\cdots\wedge\textsc{last}(u_{k})\wedge\textsc{first}(v_{k})
\end{aligned}
\right]
\end{eqnarray}

\noindent for each $i:1\le i\le k$.

Now let us construct the translation. First we define:
\begin{equation*}
\begin{aligned}
\Delta_1\,&=&\!\!\!\!\!\{&\!\!\!\!\!\!&X(\bar{z})\vee X^c(\bar{z})\,\, &\left|\right.&&\!\!\!\!\! X\in\sigma\cup\tau\,&\!\!\!\!\!\!&\left\}\right.,\\
\Delta_2\,&=&\!\!\!\!\!\{&\!\!\!\!\!\!&\textsc{Last}(\bar{x})\wedge D(\bar{x})\rightarrow X^c(\bar{z})\,\, &\left|\right.&&\!\!\!\!\! X\in\sigma&\!\!\!\!\!\!&\left\}\right.,\\
\Delta_3\,&=&\!\!\!\!\!\{&\!\!\!\!\!\!&\textsc{Last}(\bar{x})\wedge D(\bar{x})\rightarrow X(\bar{z})\,\, &\left|\right.&&\!\!\!\!\! X\in\sigma&\!\!\!\!\!\!&\left\}\right.,\\
\Delta_4\,&=&\!\!\!\!\!\left.\right\{&\!\!\!\!\!\!&\textsc{First}(\bar{x})\wedge\vartheta^c_i(\bar{x},\bar{y})\rightarrow D(\bar{x})\,\,&\left|\right.&&\!\!\!\!\!1\le i\le m&\!\!\!\!\!\!&\left\}\right.,\\
\Delta_5\,&=&\!\!\!\!\!\left.\begin{aligned}\mbox{}\\\mbox{}\end{aligned}\right\{&\!\!\!\!\!\!&
\begin{aligned}
\textsc{Succ}_j(\bar{v},\bar{x})&\wedge D(\bar{v})\wedge\\
&\vartheta^c_i(\bar{x},\bar{y})\rightarrow D(\bar{x})
\end{aligned}\,\, &\left|\begin{aligned}\mbox{}\!\\\mbox{}\!\end{aligned}\right.&&\!\!\!\!\!\!\begin{aligned}
1&\le i\le m\\1&\le j\le n\end{aligned}&\!\!\!\!\!\!&\left\}\hspace{0.005in},\begin{aligned}\,\!\!\\\,\!\!\end{aligned}\right.\\
\Delta_6\,&=&\!\!\!\!\!\{&\!\!\!\!\!\!&\textsc{Last}(\bar{x})\wedge
\neg D(\bar{x})\rightarrow\bot\,\,&&&&\!\!\!\!\!\!&\left\}\right.,
\end{aligned}
\end{equation*}
%
where, for each $X\in\sigma\cup\tau$, $X^c$ is a new predicate of the same arity; $\vartheta^c_i$ is the formula obtained from $\vartheta_i$ by substituting $X^c$ for $\neg X$ whenever $X\in\sigma\cup\tau$; $D$ is an $n$-ary new predicate.


Let $\Pi$ be the union of $\Delta_1,\dots,\Delta_6$ and $\pi$ the set of new predicates introduced in the translation. Clearly, $\exists\pi\mathrm{SM}(\Pi)$ is in $\mathrm{DLP}^n$. By a similar (slightly more complicated) argument to that in Theorem 6.3 of~\cite{EGM:97}, we can show that $\exists\pi\exists\tau\exists\sigma\mathrm{SM}(\Pi)\equiv_{\mathsf{SUC}}\exists\tau\forall\sigma\varphi$. 
%
%
%
%
\end{proof}

Next, we show that this result can be generalized to finite structures. To do this, we need a program to define the class of successor structures. Now we define it as follows.

Let $\Pi_{\textsc{s}}$ be the program consisting of the following rules.
\begin{align}
\label{eqn:succ_1}\!\!\neg\underline{\textsc{less}}(x,y)&\rightarrow\textsc{less}(x,y)\\
\label{eqn:succ_2}\!\!\neg\textsc{less}(x,y)&\rightarrow\underline{\textsc{less}}(x,y)\\
\label{eqn:succ_3}\!\!\textsc{less}(x,y)\wedge\textsc{less}(y,z)&\rightarrow\textsc{less}(x,z)\\
\label{eqn:succ_4}\!\!\textsc{less}(x,y)\wedge\textsc{less}(y,x)&\rightarrow\bot\\
\label{eqn:succ_5}\!\!\neg\textsc{less}(x,y)\wedge\neg\textsc{less}(y,x)\wedge\neg x=y&\rightarrow\bot\\
\label{eqn:succ_6}\!\!\textsc{less}(x,y)&\rightarrow\underline{\textsc{first}}(y)\\
\label{eqn:succ_7}\!\!\neg\underline{\textsc{first}}(x)&\rightarrow\textsc{first}(x)\\
\label{eqn:succ_8}\!\!\textsc{less}(x,y)&\rightarrow\underline{\textsc{last}}(x)\\
\label{eqn:succ_9}\!\!\neg\underline{\textsc{last}}(x)&\rightarrow\textsc{last}(x)\\
\label{eqn:succ_10}\textsc{less}(x,y)\wedge\textsc{less}(y,z)&\rightarrow\underline{\textsc{succ}}(x,z)\\
\label{eqn:succ_11}\neg\underline{\textsc{succ}}(x,y)\wedge\textsc{less}(x,y)&\rightarrow\textsc{succ}(x,y)
\end{align}
Informally, rules (\ref{eqn:succ_1})--(\ref{eqn:succ_2}) are choice rules to guess a binary relation $\textsc{less}$; rule (\ref{eqn:succ_3}), (\ref{eqn:succ_4}) and (\ref{eqn:succ_5}) restrict $\textsc{less}$ to be transitive, antisymmetric and total respectively so that it is a strict total order; rules (\ref{eqn:succ_6})--(\ref{eqn:succ_7}) and rules assert that $\textsc{first}$ and $\textsc{last}$ consist of the least and the last elements respectively if they exist; the last two rules then assert that $\textsc{succ}$ defines the relation for direct successors.
The following simple lemma shows that $\Pi_{\textsc{s}}$ is the desired program.


\begin{lem}\label{lem:successor}
Given a vocabulary $\sigma\supseteq\upsilon_{\textsc{s}}$ and a structure $\mathds{A}$ of $\sigma$, $\mathds{A}$ is a successor structure iff it is finite and is a model of $\exists\tau\mathrm{SM}(\Pi_{\textsc{s}})$, where $\tau$ denotes  $\upsilon(\Pi_{\textsc{s}})-\upsilon_{\textsc{s}}$.
\end{lem}


Now we can then prove the following result:

\begin{prop}\label{prop:so2dlp}
$\Sigma^{1}_{2,n}[\forall^n\exists^{\ast}]\leq_{\mathsf{FIN}}\mathrm{DLP}_n$ for all $n>1$.
\end{prop}


\begin{proof}
Let $n>1$ and $\varphi$ a sentence in $\Sigma^{1}_{2,n}[\forall^n\exists^{\ast}]$. Let $\Pi_0$ be the disjunctive program constructed in the proof of Lemma \ref{lem:so2dlp} related to $\varphi$, and let $\sigma$ be the set of predicates appearing in $\Pi_0$ but neither in $\upsilon_{\textsc{s}}$ nor in $\upsilon(\varphi)$. Let $\Pi=\Pi_0\cup\Pi_{\textsc{s}}$ and let $\tau$ be the set of predicates appearing in $\Pi_{\textsc{s}}$ but not in $\upsilon_{\textsc{s}}$.
Next we show that $\varphi$ is equivalent to
$\exists\tau\exists\sigma\mathrm{SM}(\Pi)$ over finite structures. By definition and the splitting lemma in~\cite{FLLP:09}, it suffices to show that $\varphi$ is equivalent to
\begin{equation}\label{eqn:dlp2so_fin_formula}
\exists\upsilon_{\textsc{s}}(\exists\tau\mathrm{SM}(\Pi_{\textsc{s}})\wedge\exists\sigma\mathrm{SM}(\Pi_{0}))
\end{equation}
over finite structures. Let $\upsilon$ denote the union of $\upsilon(\varphi)$ and $\upsilon_{\textsc{s}}$. Now we prove the new statement as follows.

``$\Longrightarrow$": Let $\mathds{A}$ be a finite model of $\varphi$. Clearly, there must exist at least one $\upsilon$-expansion, say $\mathds{B}$, of $\mathds{A}$ such that $\mathds{B}$ is a successor structure. By Lemma \ref{lem:successor}, $\mathds{B}$ should be a model of $\exists\tau\mathrm{SM}(\Pi_{\textsc{s}})$, and by the proof of Lemma \ref{lem:so2dlp}, $\mathds{B}$ is also a model of $\exists\sigma\mathrm{SM}(\Pi_0)$. Hence, $\mathds{A}$ is a model of formula (\ref{eqn:dlp2so_fin_formula}).

``$\Longleftarrow$": Let $\mathds{A}$ be a finite model of formula (\ref{eqn:dlp2so_fin_formula}). Then there is an $\upsilon$-expansion, say $\mathds{B}$, of $\mathds{A}$ such that $\mathds{B}$ satisfies both $\exists\tau\mathrm{SM}(\Pi_{\textsc{s}})$ and $\exists\sigma\mathrm{SM}(\Pi_0)$. By Lemma \ref{lem:successor}, $\mathds{B}$ is a successor structure, and then by the proof of Lemma \ref{lem:so2dlp}, $\mathds{B}$ must be a model of $\varphi$. This means that $\mathds{A}$ is a model of $\varphi$.
\end{proof}


With these results, we can prove a weaker separation:

\begin{thm}
$\mathrm{DLP}_{n}\not\leq_{\mathsf{FIN}}\mathrm{NLP}_{2n-1}^{\textsc{f}}$ for all $n>1$.
\end{thm}
%

\begin{proof}
Let $\upsilon_n$ be the vocabulary consisting of only an $n$-ary predicate $P_n$. Define $\textsc{Parity}^n$ to be the class of finite $\upsilon_n$-structures in each of which $P_n$ is interpreted as a set consisting of an even number of $n$-tuples. Fix $n>1$. Now, let us show that $\textsc{Parity}${\small$^{2n}$} is definable in $\mathrm{DLP}_n$ over $\mathsf{FIN}$.

We first show that, over successor structures, $\textsc{Parity}${\small$^{2n}$} is definable in $\Sigma${\small$^1_{2,n}$}$[\forall^{n}\exists^{\ast}]$. Let $\textsc{First},\textsc{Last}$ and $\textsc{Succ}_i$ be the same as those in the proof of Lemma \ref{lem:so2dlp}, and let $\textsc{Succ}(\bar{s},\bar{t})$ denote the conjunction of $\textsc{Succ}_i(\bar{s},\bar{t})$ for all $i\!:\!1\le i\le n$ if $\bar{s}$ and $\bar{t}$ are $n$-tuples of terms. Let $X$ and $Y$ be two predicate variables of arity $n$.
We define $\varphi_1$ to be the formula
\begin{equation*}
\begin{aligned}
\left[\begin{aligned}
&\forall\bar{z}(\textsc{First}(\bar{z})\rightarrow[Y(\bar{z})\leftrightarrow\,P_{2n}(\bar{x},\bar{z})])\wedge\\
&\forall\bar{y}\bar{z}(\textsc{Succ}(\bar{y},\bar{z})\rightarrow
[P_{2n}(\bar{x},\bar{z})\leftrightarrow Y(\bar{y})\oplus Y(\bar{z})])
\end{aligned}\right]\qquad\\
\rightarrow\exists\bar{z}(\textsc{Last}(\bar{z})\wedge[X(\bar{x})\leftrightarrow Y(\bar{z})]),
\end{aligned}
\end{equation*}
where $\psi\oplus\chi$ denotes the formula $(\psi\leftrightarrow\neg\chi)$. Informally, $\varphi_1$ is intended to define ``$X(\bar{a})$ is true if and only if the cardinality of $\{\bar{b}\mid P(\bar{a},\bar{b})\}$ is odd". Define $\varphi_2$ to be the formula
\begin{equation*}
\begin{aligned}
\left[\begin{aligned}
&\forall\bar{z}(\textsc{First}(\bar{z})\rightarrow[X(\bar{z})\leftrightarrow\,Y(\bar{z})])\wedge\\
&\forall\bar{y}\bar{z}(\textsc{Succ}(\bar{y},\bar{z})\rightarrow[X(\bar{z})\leftrightarrow Y(\bar{y})\oplus Y(\bar{z})])
\end{aligned}\right]\qquad\\
\rightarrow\exists\bar{z}[\textsc{Last}(\bar{z})\wedge\neg Y(\bar{z})].
\end{aligned}
\end{equation*}
Intuitively, $\varphi_2$ asserts ``$X$ consists of an even number of $n$-tuples on the domain". Now, let $\varphi=\exists X\forall Y\forall\bar{x}(\varphi_1\wedge\varphi_2)$. Obviously, $\varphi$ can be written as an equivalent sentence in $\Sigma${\small$^1_{2,n}$}$[\forall^{n}\exists^{\ast}]$. By a careful check,
it is not difficult to show that, given any successor structure $\mathds{A}$ of $\upsilon(\varphi)$, the restriction of $\mathds{A}$ to $\upsilon${\small$_{2n}$} is in $\textsc{Parity}${\small$^{2n}$} if and only if $\mathds{A}$ is a model of $\varphi$.

According to the proof of Lemma \ref{lem:so2dlp}, there exist a disjunctive program $\Pi_{0}$ and a finite set $\tau$ of predicates of arities $\le n$ such that $\exists\tau\mathrm{SM}(\Pi_0)$ is equivalent to $\varphi$ over successor structures and no predicate in $\upsilon_{\textsc{s}}$ is intensional w.r.t. $\Pi_0$. Let $\Pi$ be the union of $\Pi_{\textsc{s}}$ and $\Pi_{0}$. Let $\sigma$ be the set of predicates in $\upsilon(\Pi)-\upsilon${\small$_{2n}$}. It is easy to show that, over finite structures, $\textsc{Parity}${\small$^{2n}$} is defined by $\exists\sigma\mathrm{SM}(\Pi)$, so definable in $\mathrm{DLP}_n$.

Next, we show that $\textsc{Parity}${\small$^{2n}$} is undefinable in $\mathrm{NLP}${\small$_{2n-1}^{\textsc{f}}$} over finite structures. If this is true, we then obtain the desired proposition immediately.
By Proposition \ref{prop:nlp2eso}, it is sufficient to prove that $\textsc{Parity}${\small$^{2n}$} is not definable in $\Sigma${\small$_{1,2n-1}^{1\textsc{f}}$} over finite structures. Towards a contradiction, assume that it is not true. By a similar argument to that in Theorem 3.1 of~\cite{DLS:98}, we have:

\medskip
\noindent{\em Claim}. Let $m\ge\!1$. Then $\textsc{Parity}${\small$^{2m}$} is definable in $\Sigma${\small$_{1,2m-2}^{1\textsc{f}}$} over $\mathsf{FIN}$ if $\textsc{Parity}${\small$^{m}$} is definable in $\Sigma${\small$_{1,m-1}^{1\textsc{f}}$} over $\mathsf{FIN}$. 
\medskip

With this claim, we can then infer that $\textsc{Parity}${\small$^{4n}$} is definable in $\Sigma${\small$_{1,4n-2}^{1\textsc{f}}$} over finite structures. As every function variable of arity $k\ge 0$ can always be simulated by a predicate variable of arity $k+1$, $\textsc{Parity}${\small$^{4n}$} should be definable in $\Sigma${\small$_{1,4n-1}^{1}$} over finite structures, which contradicts with Theorem 2.1 in~\cite{Ajtai:83}. This completes the proof.
\end{proof}



\section{Arbitrary Structures}

Based on the results presented in the previous two sections, we can then compare the expressiveness of disjunctive programs and normal programs over arbitrary structures.


\begin{thm}\label{thm:arb_dlp2nlp}
$\mathrm{DLP}\simeq\mathrm{NLP}$ iff $\mathrm{DLP}\simeq_{\mathsf{FIN}}\mathrm{NLP}$.
\end{thm}

\begin{proof} 
The left-to-right direction is trivial. Now let us show the converse. Assume $\mathrm{DLP}\simeq_{\mathsf{FIN}}\mathrm{NLP}$, and let $\Pi$ be a disjunctive program. Then there must exist a normal program $\Pi^{\circ}$ such that $\mathrm{SM}(\Pi)\equiv_{\mathsf{FIN}}\exists\sigma\mathrm{SM}(\Pi^{\circ})$, where $\sigma$ is the set of predicates occurring in $\Pi^{\circ}$ but not in $\Pi$. By Theorem \ref{thm:tran_infinite}, there is a normal program $\Pi^{\diamond}$ such that $\mathrm{SM}(\Pi)\equiv_{\mathsf{INF}}\exists\tau\mathrm{SM}(\Pi^{\diamond})$. Without loss of generality, let us assume  $\sigma\cap\tau=\emptyset$. To show $\mathrm{DLP}\simeq\mathrm{NLP}$, our idea is to design a normal program testing whether or not the intended structure is finite. If that is true, we let $\Pi^{\circ}$ work; otherwise, let $\Pi^{\diamond}$ work. To do this, we introduce a new predicate $\textsc{finite}$ of arity 0, and let
%
$\Pi_{\textsc{t}}$ be the union of $\Pi_{\textsc{s}}$ and the following logic program: 
\begin{equation}
\left\{\qquad\begin{aligned}
\textsc{first}(x)&\rightarrow\textsc{num}(x),\\
\textsc{num}(x)\wedge\textsc{succ}(x,y)&\rightarrow\textsc{num}(y),\\
\textsc{num}(x)\wedge\textsc{last}(x)&\rightarrow\textsc{finite}
\end{aligned}\qquad\right\}.
\end{equation}
Let $\pi=\upsilon(\Pi_{\textsc{t}})-\{\textsc{finite}\}$. We then have the following:

\smallskip
\noindent{\em Claim.} If $\mathds{A}\models\!\exists\pi\mathrm{SM}(\Pi_{\textsc{t}})$, then $\mathds{A}$ is finite iff $\mathds{A}\models\textsc{finite}$.
\smallskip

The left-to-right direction follows from Lemma \ref{lem:successor}. We only show the converse. Let us assume that $\mathds{A}$ satisfies both $\textsc{finite}$ and $\exists\pi\mathrm{SM}(\Pi_{\textsc{t}})$. Let $\upsilon_0$ be the union of $\upsilon(\Pi_{\textsc{t}})$ and the vocabulary of $\mathds{A}$. Then, there must exist an $\upsilon_0$-expansion $\mathds{B}$ of $\mathds{A}$ such that $\mathds{B}$ is a stable model of $\Pi_{\textsc{t}}$. So, $\textsc{less}^{\mathds{B}}$ should be a strict total order on $A$; the element in $\textsc{first}^{\mathds{B}}$ (respectively, $\textsc{last}^{\mathds{B}}$), if it exists, should be the least (respectively, largest) element in $A$ w.r.t. $\textsc{less}^{\mathds{B}}$; and $\textsc{succ}^{\mathds{B}}$ should be the relation defining the direct successors w.r.t. $\textsc{less}^{\mathds{B}}$. As $\textsc{finite}$ is true in $\mathds{A}$, there must exist an integer $n\ge0$ and $n$ elements $a_1,\dots,a_n$ in $A$ such that $\textsc{first}(a_1),\textsc{last}(a_n)$ and each $\textsc{succ}(a_i,a_{i+1})$ are true in $\mathds{B}$. We assert that every element in $A$ should be $a_i$ for some $i$. If not, let $b$ be one of such elements. As $\textsc{less}^{\mathds{B}}$ is a strict total order, there must exist $i\hspace{-.03cm}:\hspace{-.03cm}1\hspace{-.03cm}\le\hspace{-.03cm}i\hspace{-.03cm}<\hspace{-.03cm}n$ such that both $\textsc{less}(a_i,b)$ and $\textsc{less}(b,a_{i+1})$ are true in $\mathds{B}$. But this is impossible since $\textsc{succ}(a_i,a_{i+1})$ is true in ${\mathds{B}}$. So, we must have $A=\{a_1,\dots,a_n\}$. This implies that $\mathds{A}$ is finite, and then we obtain the claim.

\smallskip

Next, let us construct the desired program. Let $\Pi^{\circ}_0$ (respectively $\Pi^{\diamond}_0$) denote the normal program obtained from $\Pi^{\circ}$ (respectively, $\Pi^{\diamond}$) by adding $\textsc{finite}$ (respectively, $\neg\textsc{finite}$) to the body of each rule as a conjunct. Let $\Pi^{\dag}$ be the union of $\Pi^{\circ}_0$, $\Pi^{\diamond}_0$ and $\Pi_{\textsc{t}}$. Let $\nu=\upsilon(\Pi^{\dag})-\upsilon(\Pi)$. Now, we show that $\exists\nu\mathrm{SM}(\Pi^{\dag})$ is equivalent to $\mathrm{SM}(\Pi)$ over arbitrary structures.
By definition and the splitting lemma in \cite{FLLP:09}, it suffices to show that $\mathrm{SM}(\Pi)$ is equivalent to
\begin{equation}
\exists\textsc{finite}[\exists\sigma\mathrm{SM}(\Pi^{\circ}_0)\wedge\exists\tau\mathrm{SM}(\Pi^{\diamond}_0)\wedge\exists\pi\mathrm{SM}(\Pi_{\textsc{t}})].
\end{equation}

Let $\mathds{A}$ be a structure of $\upsilon(\Pi)$. As a strict partial order always exists on domain $A$, we can construct an $\upsilon(\Pi)\cup\upsilon(\Pi_{\textsc{t}})$-expansion $\mathds{B}$ of $\mathds{A}$ such that $\mathds{B}$ is a stable model of $\Pi_{\textsc{t}}$. By the claim, $\mathds{B}\models\textsc{finite}$ if and only if $\mathds{A}$ is finite. First assume that $\mathds{A}$ is finite. By definition, it is clear that $\exists\sigma\mathrm{SM}(\Pi^{\circ}_0)$ is satisfied by $\mathds{B}$ if and only if $\exists\sigma\mathrm{SM}(\Pi^{\circ})$ is satisfied by $\mathds{A}$, and $\exists\sigma\mathrm{SM}(\Pi^{\diamond}_0)$ is always true in $\mathds{B}$. This means that $\exists\nu\mathrm{SM}(\Pi^{\dag})$ is equivalent to $\mathrm{SM}(\Pi)$ over finite structures. By a symmetrical argument, we can show that the equivalence also holds over infinite structures. This then completes the proof.
\end{proof}

\begin{rem}
In classical logic, it is well-known that separating languages over arbitrary structures is usually easier than that over finite structures~\cite{EF:99}. In logic programming, it also seems that arbitrary structures are better-behaved than finite structures. For example, there are some preservation theorems that work on arbitrary structures, but not on finite structures~\cite{AG:94}. Thus, it might be possible to develop techniques on arbitrary structures for some stronger separations of $\mathrm{DLP}$ from $\mathrm{NLP}$.
\end{rem}


\begin{cor}
$\mathrm{DLP}\simeq\mathrm{NLP}$ iff $\mathrm{NP}=\mathrm{coNP}$.
\end{cor}


Next, we give a characterization for disjunctive programs.

\begin{prop}
$\mathrm{DLP}\simeq\Sigma^{1}_{2}[\forall^{\ast}\exists^{\ast}]$.
\end{prop}

\begin{proof}
(Sketch) The direction ``$\le$" trivially follows from the second-order definition of stable model semantics. So, it remains to show the converse.
Let $\varphi$ be a sentence in $\Sigma${\small$^{1}_{2}$}$[\forall^{\ast}\exists^{\ast}]$. 
Without loss of generality, assume that $\varphi$ is of the form
\begin{equation}
\exists\tau\forall\sigma\forall\bar{x}\exists\bar{y}[\vartheta_1(\bar{x},\bar{y})\vee\cdots\vee\vartheta_k(\bar{x},\bar{y})]
\end{equation}
where $\tau$ and $\sigma$ are two finite sets of predicates; $\bar{x}$ and $\bar{y}$ two finite tuples of individual variables; each $\vartheta_i$ is a conjunction of atoms or negated atoms. Let $n$ be the length of $\bar{x}$.

Now, we construct a translation. Firstly, let us define
\begin{equation}
\begin{aligned}
\Lambda_1\,\,&=&\!\!\{&\!\!\!\!\!\!&\,T_X(\bar{x},\bar{z})\vee F_X(\bar{x},\bar{z})\,\, &\mid\,\, X\in\sigma\cup\tau\,&\!\!\!\!\!\!\}&,\\
\Lambda_2\,\,&=&\!\!\{&\!\!\!\!\!\!&D(\bar{x})\rightarrow F_X(\bar{x},\bar{z})\,\, &\mid\,\, X\in\sigma&\!\!\!\!\!\!\}&,\\
\Lambda_3\,\,&=&\!\!\{&\!\!\!\!\!\!&D(\bar{x})\rightarrow T_X(\bar{x},\bar{z})\,\, &\mid\,\, X\in\sigma&\!\!\!\!\!\!\}&,\\
\Lambda_4\,\,&=&\!\!\{&\!\!\!\!\!\!&\vartheta^{\diamond}_i(\bar{x},\bar{y})\rightarrow D(\bar{x})\,\, &\mid\,\, 1\le i\le k&\!\!\!\!\!\!\}&,\\
\Lambda_5\,\,&=&\!\!\{&\!\!\!\!\!\!&\neg D(\bar{x})\rightarrow\bot\,\,&&\!\!\!\!\!\!\}&,
\end{aligned}
\end{equation}
where, for each $X\in\sigma\cup\tau$, $T_X$ and $F_X$ are two distinct new predicates of arity $(m+n)$ if $m$ is the arity of $X$; each $\vartheta^{\diamond}_i$ is the formula obtained from $\vartheta_i$ by substituting $F_X(\bar{x},\bar{t})$ for $\neg X(\bar{t})$ and followed by substituting $T_X(\bar{x},\bar{t})$ for $X(\bar{t})$ whenever $X\in\sigma\cup\tau$ and $\bar{t}$ is a tuple of terms of the corresponding length; and $D$ is an $n$-ary new predicate.

Let $\Pi$ be the union of $\Lambda_1,\dots,\Lambda_5$. Clearly, $\Pi$ is a disjunctive program. Let $\pi$ be the set of new predicates introduced in the translation. By a similar argument to that in Lemma \ref{lem:so2dlp}, we can show that $\varphi$ is equivalent to $\exists\pi\mathrm{SM}(\Pi)$.
\end{proof}

\section{Conclusion and Related Work}

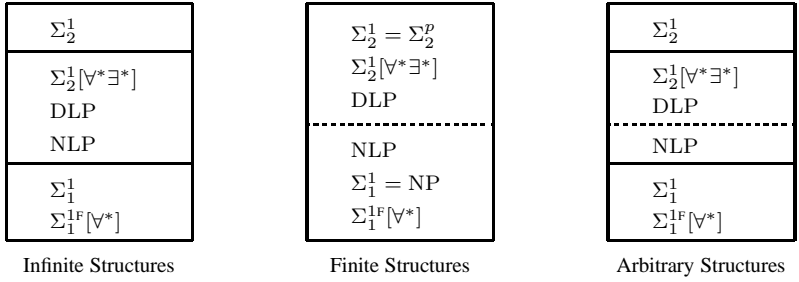
\begin{figure}[!h]
\setlength{\unitlength}{0.115in}
\begin{minipage}[t]{0.33\linewidth}
\begin{center}
\begin{picture}(8.5,11)\thicklines
\put(0,0){\line(0,1){10.8}}
\put(8.5,0){\line(0,1){10.8}}
\put(0,0){\line(1,0){8.5}}
\put(2,0.6){{\footnotesize$\Sigma^{1\textsc{f}}_{1}[\forall^{\ast}]$}}
\put(2,2.0){{\footnotesize$\Sigma^1_{1}$}}
\put(0,3.5){\line(1,0){8.5}}
\put(2,4.1){{\footnotesize$\mathrm{NLP}$}}
\put(2,5.6){{\footnotesize$\mathrm{DLP}$}}
\put(2,7.1){{\footnotesize$\Sigma^1_2[\forall^{\ast}\exists^{\ast}]$}}
\put(0,8.6){\line(1,0){8.5}}
\put(2,9.3){{\footnotesize$\Sigma^1_2$}}
\put(0,10.8){\line(1,0){8.5}}
\end{picture}\\
{\footnotesize Infinite Structures}
\end{center}
\end{minipage}%
\begin{minipage}[t]{0.33\linewidth}
\begin{center}
\begin{picture}(8.5,11)\thicklines
\put(0,0){\line(0,1){10.8}}
\put(8.5,0){\line(0,1){10.8}}
\put(0,0){\line(1,0){8.5}}
\put(2,0.8){{\footnotesize$\Sigma^{1\textsc{f}}_{1}[\forall^{\ast}]$}}
\put(2,2.3){{\footnotesize$\Sigma^1_{1}=\mathrm{NP}$}}
\put(2,3.8){{\footnotesize$\mathrm{NLP}$}}
\put(0,5.3){\dashbox{0.2}(8.5,0){}}
\put(2,6.1){{\footnotesize$\mathrm{DLP}$}}
\put(2,7.6){{\footnotesize$\Sigma^1_2[\forall^{\ast}\exists^{\ast}]$}}
\put(2,9.1){{\footnotesize$\Sigma^1_2=\Sigma^p_2$}}
\put(0,10.8){\line(1,0){8.5}}
\end{picture}\\
{\footnotesize Finite Structures}
\end{center}
\end{minipage}%
\begin{minipage}[t]{0.33\linewidth}
\begin{center}
\begin{picture}(8.5,11)\thicklines
\put(0,0){\line(0,1){10.8}}
\put(8.5,0){\line(0,1){10.8}}
\put(0,0){\line(1,0){8.5}}
\put(2,0.6){{\footnotesize$\Sigma^{1\textsc{f}}_{1}[\forall^{\ast}]$}}
\put(2,2.0){{\footnotesize$\Sigma^1_{1}$}}
\put(0,3.5){\line(1,0){8.5}}
\put(2,4.0){{\footnotesize$\mathrm{NLP}$}}
\put(0,5.3){\dashbox{0.2}(8.5,0){}}
\put(2,5.8){{\footnotesize$\mathrm{DLP}$}}
\put(2,7.2){{\footnotesize$\Sigma^1_2[\forall^{\ast}\exists^{\ast}]$}}
\put(0,8.6){\line(1,0){8.5}}
\put(2,9.3){{\footnotesize$\Sigma^1_2$}}
\put(0,10.8){\line(1,0){8.5}}
\end{picture}\\
{\footnotesize Arbitrary Structures}
\end{center}
\end{minipage}
\caption{Expressiveness Hierarchies Related to LPs}\label{figure:conclusion}
\end{figure}

Combining the results proved in previous sections with some existing results, we then obtain an almost complete picture for the expressiveness of logic programs and some related fragments of second-order logic. As shown in Figure \ref{figure:conclusion}, the expressiveness hierarchy in each subfigure is related to a structure class. In each subfigure, the syntactical classes in a same block are proved to be of the same expressiveness over the related structure class. A block is closer to the top, the classes in the block are then more expressive. In addition, a dashed line means that the corresponding separation is true if and only if $\mathrm{NP}$ is not closed under complement.

Without involving the well-known complexity conjecture, we established the intranslatability from disjunctive to normal programs over finite structures if the arities of auxiliary constants are bounded in a certain sense. This can be regarded as evidence that disjunctive programs are more expressive than normal programs over finite structures. As a byproduct, we also developed a succinct translation from normal programs to first-order sentences. This then improved the ordered completion proposed by~\cite{ALZZ:12}.

There are several existing works contributing to Figure \ref{figure:conclusion}, which are listed as follows. The translatability from $\Sigma${\small$^1_1$} to $\Sigma${\small$^{1\textsc{f}}_1$}$[\forall^{\ast}\hspace{-.03cm}]$ follows from the well-known Skolem normal form. The translatability from $\Sigma^1_2$ to $\Sigma${\small$^1_2$}$[\forall^{\ast}\hspace{-.03cm}\exists^{\ast}\hspace{-.03cm}]$ over finite structures is due to~\cite{Leiv:89}. The separation of $\Sigma${\small$^1_2$} from $\Sigma${\small$^1_2$}$[\forall^{\ast}\hspace{-.03cm}\exists^{\ast}\hspace{-.03cm}]$ (on both arbitrary and infinite structures) is due to~\cite{EGG:96}. From $\mathrm{NLP}$ to $\Sigma${\small$^1_1$}, both the intranslatability over arbitrary structures and the translatability over finite structures are due to~\cite{ALZZ:12}.



The (in)translatability between first-order theories and logic programs were also considered in~\cite{ZZYZ:11}. But first-order theories there are based on non-monotonic semantics. Over Herbrand structures, \cite{Schl:95,EG:97} proved that normal programs, disjunctive programs and universal second-order logic are of the same expressiveness under the query equivalence. Their proofs employ an approach from  recursion theory. However, this approach seems difficult to be applied to general infinite structures. In the propositional case, there have been a lot of works on the translatability and expressiveness of logic programs, e.g., \cite{EFTW:04,Janhunen:06}. It should be noted that the picture of expressiveness and translatability in there is quite different from that in the first-order case.


{
\bibliographystyle{amsplain}
\bibliography{kr}

\providecommand{\bysame}{\leavevmode\hbox to3em{\hrulefill}\thinspace}
\providecommand{\MR}{\relax\ifhmode\unskip\space\fi MR }
\providecommand{\MRhref}[2]{%
  \href{http://www.ams.org/mathscinet-getitem?mr=#1}{#2}
}
\providecommand{\href}[2]{#2}
\begin{thebibliography}{10}

\bibitem{Ajtai:83}
M.~Ajtai, \emph{{$\Sigma^1_1$}-formulae on finite structures}, Annals of Pure
  and Applied Logic \textbf{24} (1983), 1--48.

\bibitem{AG:94}
M.~Ajtai and Y.~Gurevich, \emph{Datalog vs first-order logic}, Journal of
  Computer and System Sciences \textbf{49} (1994), 562--588.

\bibitem{ALZZ:12}
V.~Asuncion, F.~Lin, Y.~Zhang, and Y.~Zhou, \emph{Ordered completion for
  first-order logic programs on finite structures}, Artificial Intelligence
  \textbf{177--179} (2012), 1--24.

\bibitem{BD:94}
R.~Ben-Eliyahu and R.~Dechter, \emph{Propositional semantics for disjunctive
  logic programs}, Annals of Mathematics and Artificial Intelligence
  \textbf{12} (1994), no.~1--2, 53--87.

\bibitem{DEGV:01}
E.~Dantsin, T.~Eiter, G.~Gottlob, and A.~Voronkov, \emph{Complexity and
  expressive power of logic programming}, ACM Computing Surveys \textbf{33}
  (2001), no.~3, 374--425.

\bibitem{DGO:04}
A.~Durand, E.~Grandjean, and F.~Olive, \emph{New results on arity vs. number of
  variables}, Research report 20--2004, LIF, \textbf{Marseille, France} (2004).

\bibitem{DLS:98}
A.~Durand, C.~Lautemann, and T.~Schwentick, \emph{Subclasses of binary {NP}},
  Journal of Logic and Computation \textbf{8} (1998), no.~2, 189--207.

\bibitem{EF:99}
H.-D. Ebbinghaus and J.~Flum, \emph{Finite model theory}, 2 ed.,
  Springer-Verlag, New York, 1999.

\bibitem{EG:97}
T.~Eiter and G.~Gottlob, \emph{Expressiveness of stable model semantics for
  disjunctive logic programs with functions}, The Journal of Logic Programming
  \textbf{33} (1997), 167--178.

\bibitem{EGG:96}
T.~Eiter, G.~Gottlob, and Y.~Gurevich, \emph{Normal forms for second-order
  logic over finite structures, and classication of {NP} optimization
  problems}, Annals of Pure and Applied Logic \textbf{78} (1996), 111--125.

\bibitem{EGM:97}
T.~Eiter, G.~Gottlob, and H.~Mannila, \emph{Disjunctive datalog}, ACM
  Transactions on Database Systems \textbf{22} (1997), 364--418.

\bibitem{EFTW:04}
Thomas Eiter, Michael Fink, Hans Tompits, and Stefan Woltran, \emph{On
  eliminating disjunctions in stable logic programming}, Proceedings of KR,
  2004, pp.~447--458.

\bibitem{Fagin:74}
R.~Fagin, \emph{Generalized first-order spectra and polynomial-time
  recognizable sets}, Complexity of Computation, SIAM-AMS Proceedings, vol.~7,
  1974, pp.~43--73.

\bibitem{FLL:11}
P.~Ferraris, J.~Lee, and V.~Lifschitz, \emph{Stable models and
  circumscription}, Artificial Intelligence \textbf{175} (2011), 236--263.

\bibitem{FLLP:09}
P.~Ferraris, J.~Lee, V.~Lifschitz, and R.~Palla, \emph{Symmetric splitting in
  the general theory of stable models}, Proceedings of IJCAI, 2009,
  pp.~797--803.

\bibitem{GL:88}
M.~Gelfond and V.~Lifschitz, \emph{The stable model semantics for logic
  programming}, Proceedings of ICLP/SLP, 1988, pp.~1070--1080.

\bibitem{Gran85}
Etienne Grandjean, \emph{Universal quantifiers and time complexity of random
  access machines}, Mathematical Systems Theory \textbf{18} (1985), no.~2,
  171--187.

\bibitem{Imme:99}
Neil Immerman, \emph{Descriptive complexity}, Graduate texts in computer
  science, Springer, 1999.

\bibitem{Janhunen:06}
Tomi Janhunen, \emph{Some (in)translatability results for normal logic programs
  and propositional theories}, Journal of Applied Non-Classical Logics
  \textbf{16} (2006), no.~1--2, 35--86.

\bibitem{Leiv:89}
D.~Leivant, \emph{Descriptive characterizations of computational complexity},
  Journal of Computer and System Sciences \textbf{39} (1989), 51--83.

\bibitem{LM:04}
Yuliya Lierler and Marco Maratea, \emph{Cmodels-2: {SAT}-based answer set
  solver enhanced to non-tight programs}, Proceedings of LPNMR, 2004,
  pp.~346--350.

\bibitem{LZ:04}
F.~Lin and Y.~Zhao, \emph{Assat: computing answer sets of a logic program by
  sat solvers}, Artificial Intelligence \textbf{157} (2004), no.~1-2, 115--137.

\bibitem{LZ:11}
F.~Lin and Y.~Zhou, \emph{From answer set logic programming to circumscription
  via logic of {GK}}, Artificial Intelligence \textbf{175} (2011), no.~1,
  264--277.

\bibitem{LMR:92}
J.~Lobo, J.~Minker, and A.~Rajasekar, \emph{Foundations of disjunctive logic
  programming}, The MIT Press, Cambridge, 1992.

\bibitem{PV:05}
David Pearce and Agust¨ªn Valverde, \emph{A first order nonmonotonic extension
  of constructive logic}, Studia Logica \textbf{80} (2005), no.~2/3, 321--346.

\bibitem{Schl:95}
J.~S. Schlipf, \emph{The expressive powers of the logic programming semantics},
  Journal of Computer and System Sciences \textbf{51} (1995), no.~1, 64--86.

\bibitem{Stoc:77}
L.~J. Stockmeyer, \emph{The polynomial-time hierarchy}, Theoretical Computer
  Science \textbf{3} (1977), 1--22.

\bibitem{ZZ:13}
Heng Zhang and Yan Zhang, \emph{First-order expressibility and boundedness of
  disjunctive logic programs}, Proceedings of IJCAI, 2013, pp.~1198--1204.

\bibitem{ZZYZ:11}
Heng Zhang, Yan Zhang, Mingsheng Ying, and Yi~Zhou, \emph{Translating
  first-order theories into logic programs}, IJCAI, 2011, pp.~1126--1131.

\end{thebibliography}
}

\end{document}